\documentclass[10pt,twocolumn,letterpaper]{article}

\usepackage{cvpr}              
\definecolor{cvprblue}{rgb}{0.21,0.49,0.74}
\usepackage[pagebackref,breaklinks,colorlinks,allcolors=cvprblue]{hyperref}
\usepackage[accsupp]{axessibility}  

\usepackage{booktabs}       
\usepackage{amsfonts}       
\usepackage{xcolor}         
\usepackage{kotex}
\usepackage{multirow}
\usepackage{colortbl}     
\usepackage[accsupp]{axessibility}

\newcommand{\myparagraph}[1]{\vspace{1pt}\noindent{\bf #1}}

\def\paperID{} 
\def\confName{CVPR}
\def\confYear{2026}

\title{Focus, Don’t Prune: Identifying Instruction-Relevant Regions for\\Information-Rich Image Understanding}

\author{
    Mincheol Kwon$^{1}$, Minseung Lee$^{1}$, Seonga Choi$^{1}$, Miso Choi$^{1}$, Kyeongjin Oh$^{2}$,\\
    Hyunyoung Lee$^{2}$, Cheonyoung Park$^{2}$, Yongho Song$^{2}$, Seunghyun Park$^{3, \dagger}$, Jinkyu Kim$^{1,4, \dagger}$\\[0.2cm]
    $^{1}$Korea University \quad $^{2}$KT Corporation \quad $^{3}$Soongsil University \quad $^{4}$Kakao Mobility\\
    \vspace{0.2 em}
    {\normalsize \texttt{\{kwonmc, jinkyukim\}@korea.ac.kr, sh.park@ssu.ac.kr}} \\
    { \url{https://github.com/minckwon/PinPoint}}
}

\begin{document}
\maketitle
\let\thefootnote\relax\footnotetext{$^{\dagger}$Corresponding authors.}

\begin{abstract}
Large Vision-Language Models (LVLMs) have shown strong performance across various multimodal tasks by leveraging the reasoning capabilities of Large Language Models (LLMs). However, processing visually complex and information-rich images, such as infographics or document layouts, requires these models to generate a large number of visual tokens, leading to significant computational overhead. To address this, we propose PinPoint, a novel two-stage framework that first identifies instruction-relevant image regions and then refines them to extract fine-grained visual features for improved reasoning and efficiency. Central to our approach is the Instruction-Region Alignment, which localizes relevant regions using both visual input and textual instructions. We further introduce new annotations that provide richer ground-truth supervision for instruction-relevant regions across challenging VQA benchmarks: InfographicVQA, MultiPageDocVQA, and SinglePageDocVQA. Experimental results show that PinPoint not only achieves superior accuracy compared to existing methods but also reduces computational overhead by minimizing irrelevant visual tokens. 
\end{abstract}    
\vspace{-1.25em}

\section{Introduction}
\label{sec:intro}
\vspace{-.25em}

\begin{figure}[t!]
  \centering
  \includegraphics[width=0.95\linewidth]{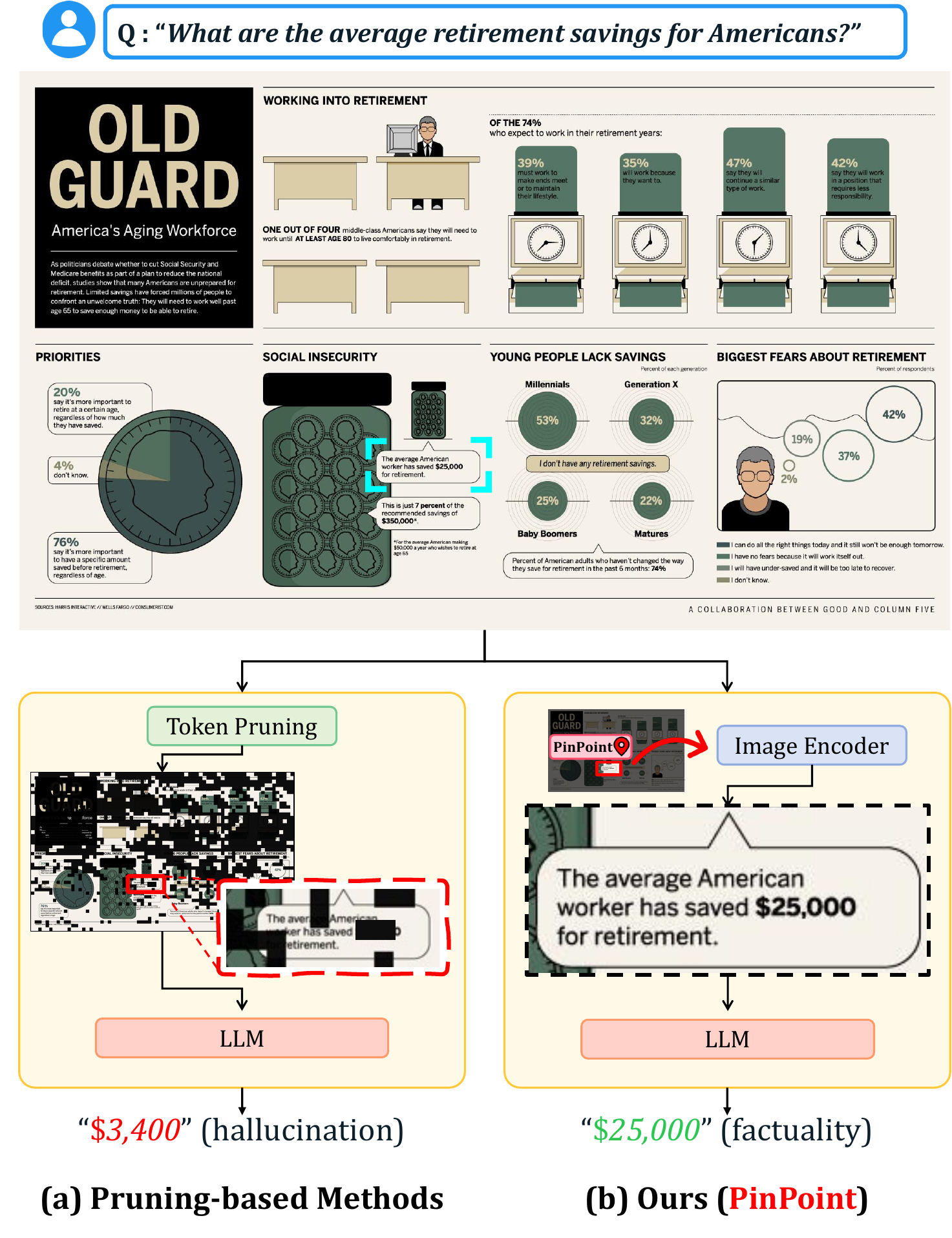}
  \caption{
  Given an input instruction (e.g., ``What are the average retirement savings for Americans?''), (a) a conventional approach aims to prune tokens based on attention weights; however, this method is often unreliable and can produce hallucinated outputs. (b) Our method, called PinPoint, first identifies instruction-relevant regions and then refines these regions to retain only the most relevant tokens, thereby improving the model’s factuality.
      }
  \vspace{-1.5em}
  \label{fig:teaser}
\end{figure}
Large Vision-Language Models (LVLMs) have achieved significant progress and strong performance across a range of multimodal downstream tasks~\cite{minigpt, videollama, videochatgpt, RegionVLM, weaklyseg}, such as visual question answering~\cite{vqa, textvqa, chartqa}, largely attributed to the increasing availability of data, enhanced computational resources, and larger model capacities. By leveraging the advanced reasoning capabilities of large language models (LLMs), recent LVLMs demonstrate superior ability in tackling complex vision-language problems. However, despite these advancements, understanding abundant fine-grained visual content, such as infographics or text-rich images (see Figure~\ref{fig:teaser}), often necessitates processing at high resolution, which in turn generates a large number of visual tokens. This approach is computationally heavy and therefore impractical.


To mitigate this computational burden, numerous token pruning methods have been proposed, typically estimating the importance of each visual token based on attention weights from the LLM's decoding layers, and subsequently removing less informative tokens to reduce computational overhead~\cite{pyramiddrop,sparsevlm,fitprune,fastv}. While these approaches show promise, three key challenges remain: (i) Imperfect attention maps — attention weights can be unreliable~\cite{pai,vissink,sparc}, sometimes leading to hallucinations, which in turn limit the effectiveness of attention-based token pruning; (ii) Semantic fragmentation — since visual elements (e.g., text) often span multiple tokens, pruning decisions at the individual token level risk disregarding their collective contextual meaning; (iii) Contextual entanglement — due to the global nature of the self-attention mechanism, visual tokens within the answer-relevant region may become entangled with irrelevant contextual information, reducing their discriminative power and effectiveness. 

\begin{figure}[t]
  \centering
  \includegraphics[width=\columnwidth]{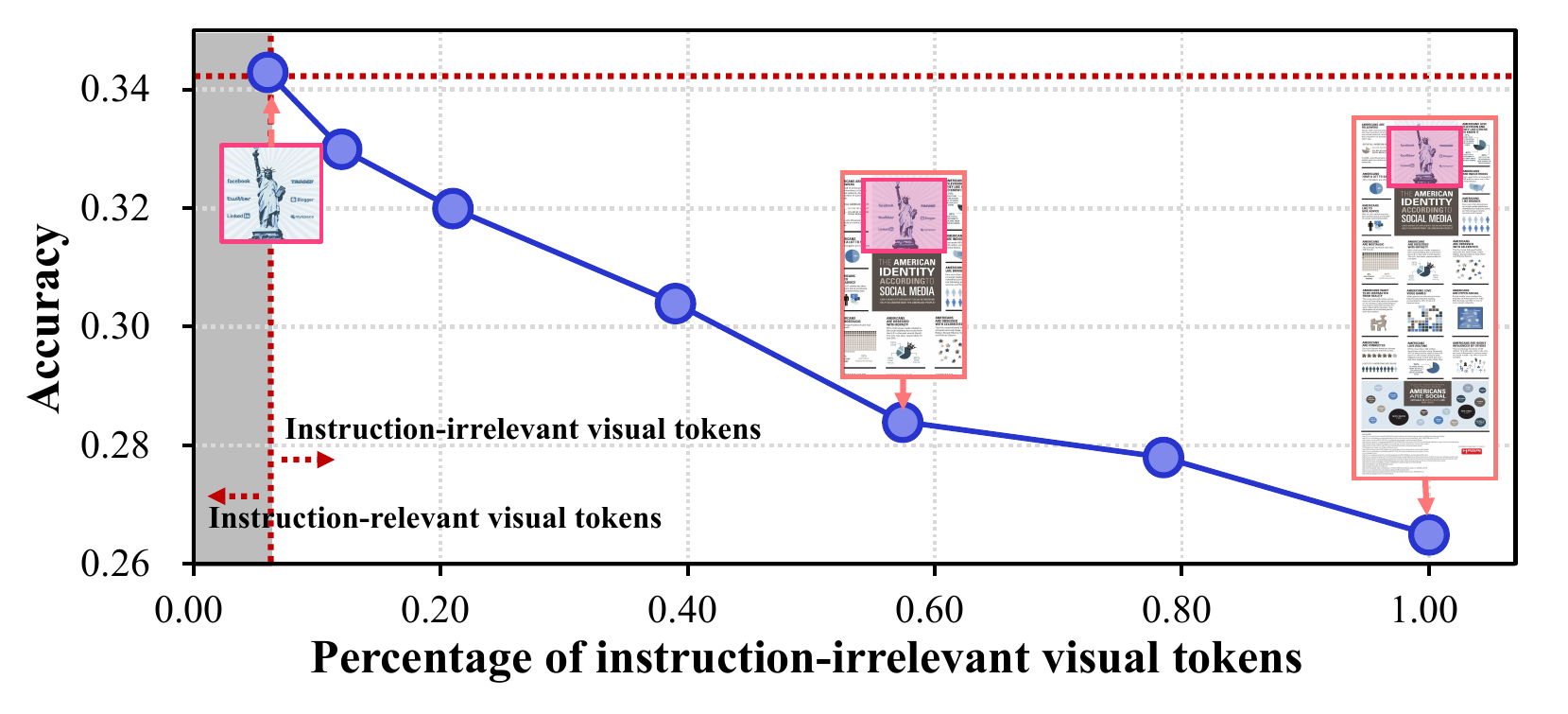}
  \caption{
  VQA performance increases with a larger proportion of instruction-relevant visual tokens. Model: LLaVA-NeXT~\cite{llavanext}. Data:  InfographicVQA~\cite{infographicvqa}.
  }
  \vspace{-1.8em}
  \label{fig:observation_1}
\end{figure}

In contrast to pruning at the individual token level, an intuitive and resource-efficient approach to understanding images rich in textual and visual content is to focus selectively on specific regions rather than processing the entire image at once. For instance, it is natural to scan the image first globally to identify regions that are likely to contain relevant information, and then concentrate on those areas in greater detail to derive an answer—similar to the strategy employed when solving tasks like "Where's Wally?"~\cite{whereswally}. Our preliminary experiments further demonstrate that LVLMs achieve improved performance when provided exclusively with question-relevant visual tokens, compared to a mixture of relevant and irrelevant tokens. As shown in Figure~\ref{fig:observation_1}, the model's performance degrades as the proportion of irrelevant tokens increases—showing the highest accuracy when using tokens from the ground-truth image region and the lowest when using tokens from all image regions indiscriminately. This selective focus can be highly efficient, as relevant regions can be processed with fewer tokens.

Thus, as illustrated in Figure~\ref{fig:teaser}(b), this paper propose a novel two-stage approach, PinPoint. This method first identifies instruction-relevant image regions and generates the fine-grained visual features required for the given instruction through visual refinement, yielding a compact set of tokens that is richer in answer-relevant information. The resulting instruction-relevant tokens are then passed to the LLM for final reasoning. This is achieved through a simple yet effective module, which employs learnable guidance queries to accurately localize instruction-relevant regions based on both visual input and textual instructions.

Our proposed method is evaluated on four challenging, information-rich VQA benchmarks: InfographicVQA (InfoVQA)~\cite{infographicvqa}, MultiPageDocVQA (MPDocVQA)~\cite{mpdocvqa}, SinglePageDocVQA (SPDocVQA)~\cite{docvqa}, and GQA~\cite{gqa}. For these datasets, ground-truth instruction-relevant regions were additionally curated. Unlike existing benchmarks~\cite{visualcot}, which typically annotate only a single bounding box containing the final answer, our annotations include multiple bounding boxes that collectively capture the necessary supporting elements. This provides richer contextual grounding to facilitate more robust reasoning and answer generation. These datasets will be made publicly available. Our contributions are as follows: 
\begin{itemize}[topsep=0pt, partopsep=0pt, itemsep=3.5pt, parsep=0pt]
    \item We propose PinPoint, a method that effectively identifies instruction-relevant regions and applies region refinement to provide richer visual information to the LLM, while significantly reducing the number of visual tokens.
    \item We provide a dataset that includes not only the ground-truth answer regions but also bounding boxes for surrounding elements necessary for answering the question, along with an automated pipeline for generating them.
    \item PinPoint outperforms existing methods in accuracy while requiring significantly fewer computational resources, as demonstrated on InfoVQA, SPDocVQA, MPDocVQA, and GQA benchmarks—and conduct experiments on LLaVA-NeXT and Qwen2-VL.
\end{itemize}
\begin{figure*}[t!]
  \centering
  \includegraphics[width=1.0\linewidth]{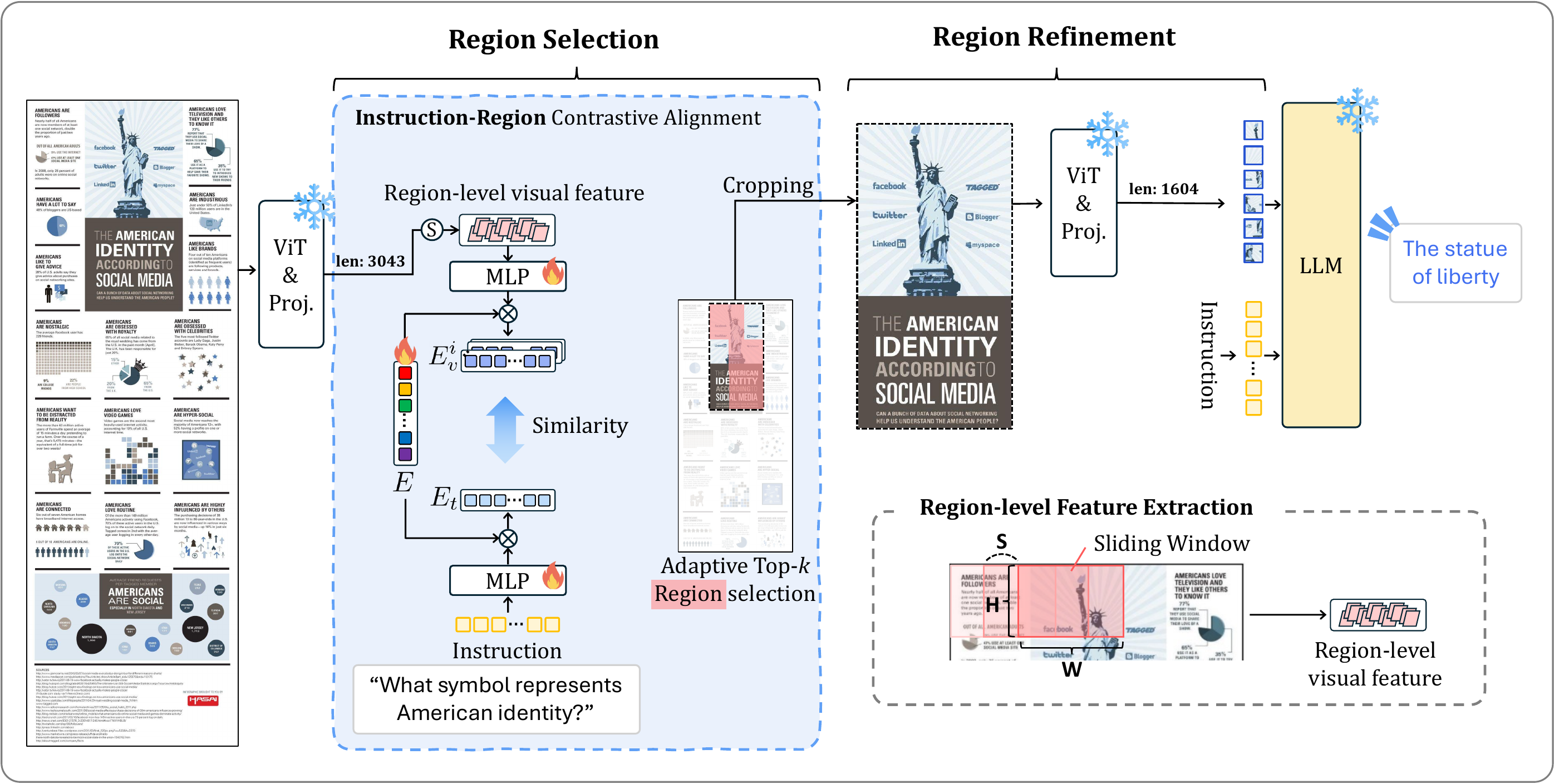}
  \caption{
  An overview of the proposed PinPoint architecture, which comprises two main stages: (1) Region Selection and (2) Region Refinement. Given a user instruction (e.g., “What symbol represents American identity?”), the Region Selection stage identifies adaptive top-$k$ instruction-relevant image regions (e.g., areas surrounding the Statue of Liberty) using region-level feature extraction via a sliding window approach. The Region Refinement stage then further processes these regions to extract fine-grained visual tokens, which are subsequently fed into a large language model (LLM) to generate the final answer (e.g., “The statue of liberty”).
  }
  \vspace{-1em}
  \label{fig:main_arch}
\end{figure*}

\section{Related Work}
\subsection{Large Vision-Language Models (LVLM)}
Recently, various LVLM models have adopted Vision Transformers (ViTs)~\cite{vit} with increased input sizes to better capture fine-grained visual information~\cite{llavanext,qwen,chen2024internvlscalingvisionfoundation}.
However, recent models have emerged to better handle high-resolution and distorted-aspect-ratio images by dividing them into ratio-preserving patches and independently applying vision encoders to each, thereby generating more visual tokens.
LLaVA-1.5~\cite{improvedllava} was proposed to handle input images up to 448×448 without resizing, using only a 224×224 visual encoder.
Subsequently, various models~\cite{monkey, internvl15, mplugdocowl15} have adopted adaptive modules that divide images into multiple patches based on pre-defined aspect ratios, enabling the processing of higher-resolution inputs. However, as the number of visual tokens increases, the computational cost grows significantly and greater reasoning capacity is required. Our proposed model, \textit{PinPoint}, focuses only on specific regions, enabling it to capture detailed information with a small number of tokens.

\subsection{Efficiency of Token Usage} 
Early efforts to improve the efficiency of Large Vision-Language Models (LVLMs) leveraged feature aggregation via learnable queries and cross-attention mechanisms~\cite{textmonkey, blip2}. However, these approaches typically aggregated visual features without explicit guidance from the given instruction. Although models such as InstructBLIP~\cite{instructblip} incorporated instruction-aware aggregation, the resulting representations often remain susceptible to noise introduced by instruction-irrelevant image regions.

Another line of work focuses on token pruning~\cite{fastv, pyramiddrop, sparsevlm, ivtp, madtp, llava-prumerge}, which aims to reduce computation by estimating token importance using attention scores from the LLM’s decoding layers and discarding low-scoring tokens—often at specific layers or across layer groups. However, in complex scenarios, attention distributions can be inaccurate or unstable, leading to hallucinations. Pruning based on such flawed attention maps risks removing semantically crucial visual tokens, thereby impairing its reasoning capabilities. This limitation is especially problematic in tasks demanding fine-grained perception, where the removal of even a single relevant visual token can significantly degrade performance.

In contrast to pruning methods that rely on potentially unstable attention maps, some approaches attempt to identify relevant regions through repeated interaction with the LLM~\cite{seal, mllmsknow}. While more precise, such methods are computationally intensive. PinPoint offers a balanced solution, using a lightweight query to efficiently identify instruction-relevant areas without the need for iterative decoding.
\section{Method}

We introduce PinPoint, a novel method designed to effectively generate instruction-relevant visual tokens. PinPoint preserves fine-grained, context-aware visual information from a large pool of tokens derived from the entire input image, thereby improving both computational efficiency and reasoning accuracy. As illustrated in Figure~\ref{fig:main_arch}, our approach consists of two key stages: (i) Region Selection, extracts coarse, region-level visual representations from the entire image. It then learns an alignment between these representations and the textual instruction to identify the most relevant regions; and (ii) Region Refinement, which extracts fine-grained visual contexts by refining the selected regions and encoding non-contextualized visual tokens. The resulting visual tokens are subsequently fed into an LLM along with the textual instruction to generate the final verdict. 
\subsection{Region-Level Feature Extraction}
\label{subsec:feature_extraction}
Accurate image understanding requires capturing contextual relationships, an aspect often missed by token-level comparisons. To address this, comparisons are performed at the region-level rather than the token-level. This approach enables us to work with more contextually rich visual features that align better with the textual instructions.

To effectively compare with the instruction information, the visual token representations are used immediately before they enter the LLM—i.e., after the pretrained Vision Transformer (ViT)~\cite{vit} and the projection layer, yielding $\mathbf{V} \in \mathbb{R}^{T \times d}$,
where $T$ is the total number of visual tokens and \( d \) is the token embedding dimension. 

Building on this representation, the visual tokens are first reshaped into a 2D spatial grid and apply a sliding window of size \( W \times H \) with stride \( S \) is applied over the token map. For \( i^{\mathrm{th}} \) window, all visual tokens within the window are collected and concatenated to form a region-level representation, \( \mathbf{R}_i \in \mathbb{R}^{W \times H \times d} \), which is the set of visual tokens in the \( i^{\mathrm{th}} \) region.

Applying the sliding window across the entire token grid yields a collection of region-level token sets, \( \mathbf{R} = \{\mathbf{R}_i \}_{i=1}^{N_r} \), where \( N_r \) is the total number of regions obtained from the sliding operation, which is input-dependent. In parallel, the input text is tokenized using byte pair encoding (BPE)~\cite{bpe} and then embedded to form the text embedding \( \mathbf{T} \in \mathbb{R}^{M \times d} \), where \( M \) is the number of text tokens.

\subsection{Instruction-Region Alignment}
\label{subsec:alignment}
Directly comparing the visual and textual features obtained above faces two fundamental challenges: first, decoder-only language models do not include a CLS token to aggregate overall semantic context; second, BPE-produced~\cite{bpe} subword units tokenization are not aligned with the visual feature space. To alleviate this problem, a module is designed to align visual and textual features, highlight query-relevant image regions, and determine which areas to focus on.

We utilize learnable guidance queries \( E = \{E_{1}, E_{2}, \ldots, E_{K}\} \in \mathbb{R}^{K \times d} \) as a common feature space to bridge visual and textual modalities, where \( K \) is the number of learnable queries. These tokens interact with both modalities to extract modality-aware representations. 
To further facilitate the alignment between two modalities, region-level visual features \( \mathbf{R}_i \) and textual features \( \mathbf{T} \) are processed by two simple MLP layers, resulting in \( \mathbf{R}_i' \) and \( \mathbf{T'} \), respectively.

To obtain visual-aware representation grounded in the shared tokens, a scaled dot-product attention is applied between \( E \) and \( i^{\mathrm{th}} \) region visual features \( \mathbf{R}_i' \), resulting in \( E^v_i \in \mathbb{R}^{K \times d}\) as follows:
\vspace{-0.75em}
\begin{align}
A_i^v = &\operatorname{softmax}\!\left(\frac{E\,\mathbf{R}_i'\,^{\!\top}}{\sqrt{d_k}}\right), \quad A^t = \operatorname{softmax}\!\left(\frac{E\,{\mathbf{T'}}\,^{\!\top}}{\sqrt{d_k}}\right) \\
E^v_i &= A^v_i \cdot \mathbf{R}_i', \qquad \qquad \quad \quad E^t = A^{t} \cdot \mathbf{T'},
\end{align}
where \( d_k \) is the scaling factor. 
Similarly, the interaction between the learnable queries \( E \) and the textual features \( \mathbf{T'} \) yields the text-aware representation \( E^t \in \mathbb{R}^{K \times d} \) through the same mechanism.

\begin{figure}[t!]
  \centering
  \includegraphics[width=\linewidth]{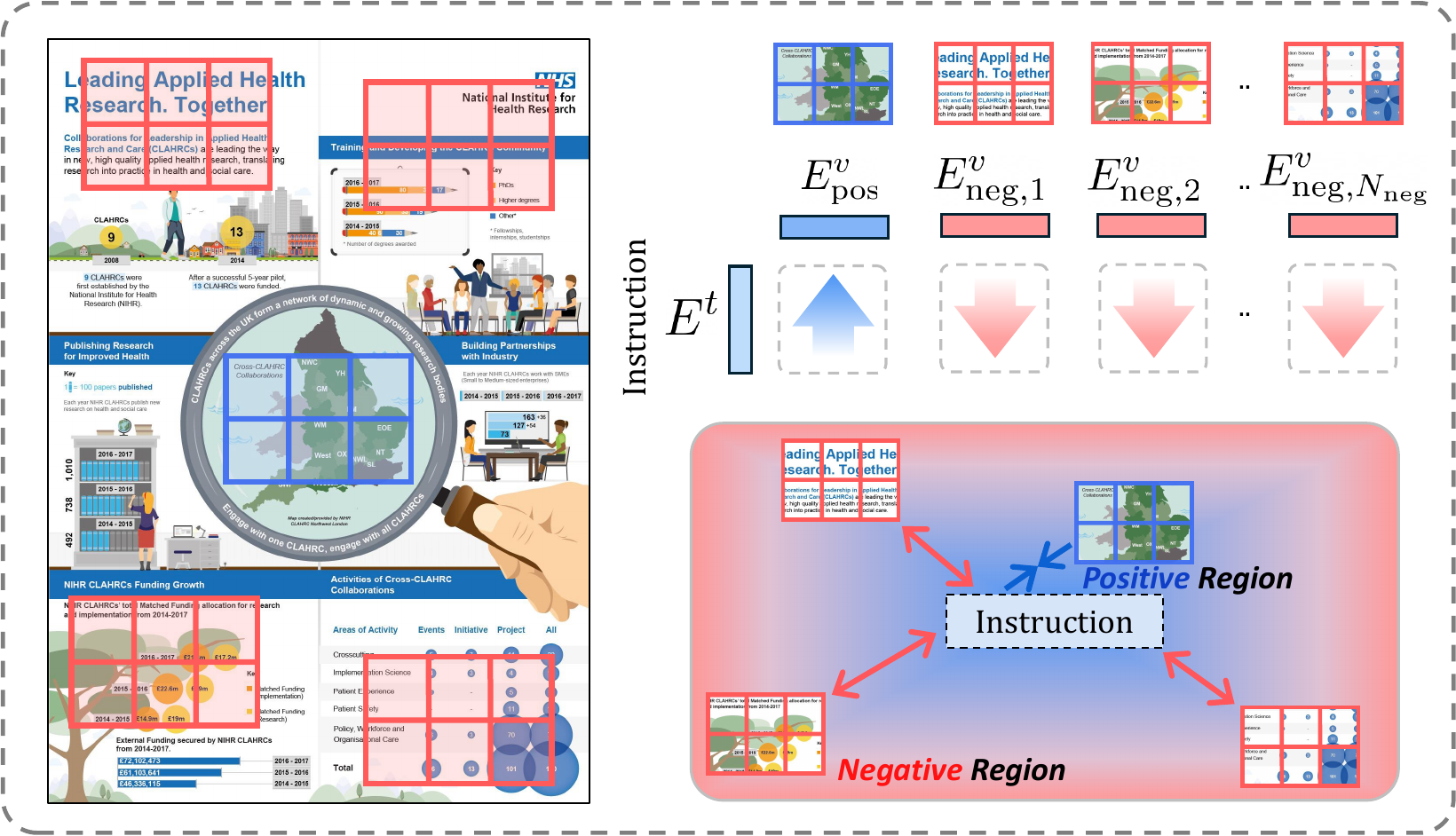}
  \caption{Illustration for intra-image contrastive loss. Region features required to answer the given instruction are treated as positives, while all other regions are treated as negatives. The loss function pulls the positive pair \( (E^t, E^v_{\text{pos}}) \) closer in the embedding space, while pushing away the negative pairs \( (E^t, E^v_{\text{neg},i}) \).}
\vspace{-1.25em}
  \label{fig:sub_figure}
\end{figure}

%
%
%
\begin{figure*}[t]
  \centering
  \includegraphics[width=\linewidth]{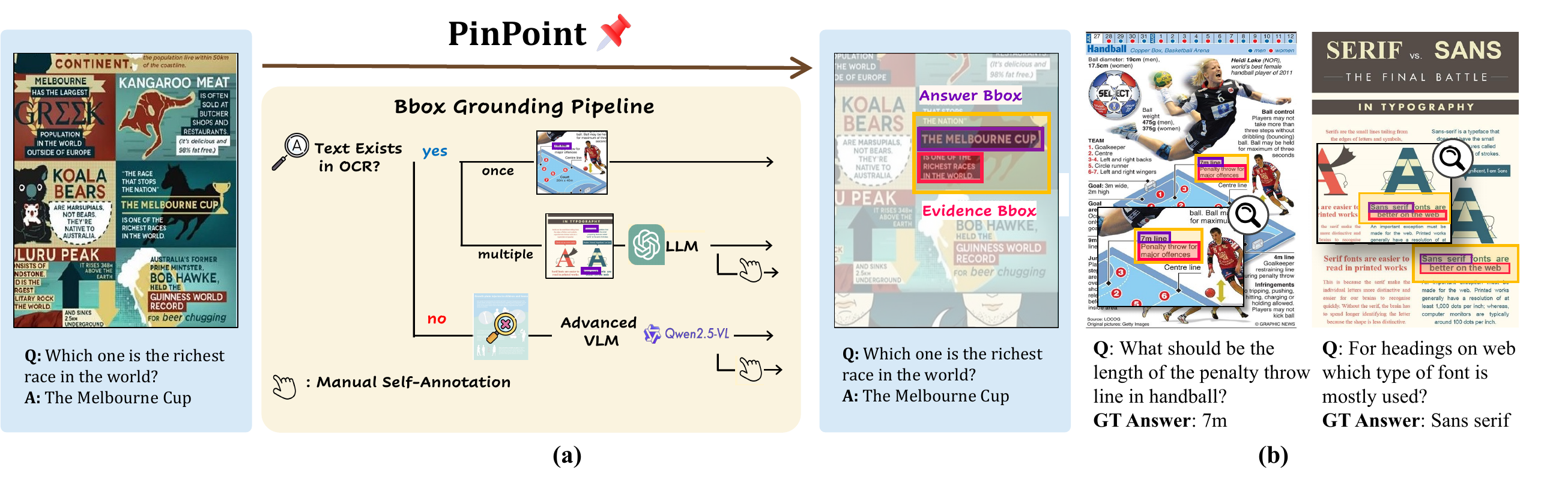}
  \vspace{-1.25em}
  \caption{
  (a) Built upon existing VQA benchmarks, such as InfoVQA~\cite{infographicvqa}, SPDocVQA~\cite{docvqa}, and MPDocVQA~\cite{mpdocvqa}, we construct a new dataset using a pipeline that provides annotations of instruction-relevant regions likely to contain answers and supporting evidence. (b) Examples of the annotated regions. More examples are available in the supplemental material.
  }
\vspace{-1.25em}
  \label{fig:dataset}
\end{figure*}
%
%
%

\subsection{Selecting and Refining Instruction-Relevant Regions}
\label{subsec:region_selection}
\paragraph{Region Selection.}
The aligned text features from Sec. \ref{subsec:alignment} are leveraged to identify instruction-relevant image regions. All candidate regions are ranked by their feature similarity to the instruction (i.e., cosine similarity between \(E^v_i\) and \(E^t\)), then the top-ranked regions are adaptively selected until their enclosing region exceeds a predefined ratio $r$ of the total image area.

\paragraph{Region Refinement.} 
The initial visual tokens from ViT~\cite{vit} are often contaminated by irrelevant image areas due to global self-attention. To address this, only the regions selected in the previous step are re-encoded. This refinement re-encodes these tokens in isolation, stripping away the irrelevant global context. The result is a denser, more informative, and significantly smaller set of visual tokens, which is both computationally efficient and more effective for downstream reasoning.

\subsection{Training with Contrastive Loss}
\label{subsec:contrastive_loss}

To enable effective alignment of visual and textual features, two contrastive learning objectives are introduced: (i) an \textbf{inter-modal contrastive loss} to align representations between modalities, and (ii) an \textbf{intra-image contrastive loss} to better separate instruction-relevant regions from irrelevant ones.

To define the samples for these losses, positive and negative regions are first identified based on the instruction and its ground truth (GT) region.
The region obtained from the sliding window that is closest to the center of the GT region is treated as the positive sample, denoted $E^v_{\text{pos}}$.
Since the GT region (especially for $\mathcal{L}_{\text{intra}}$) may span multiple regions, only those samples from answer-irrelevant regions are considered as negative samples, denoted $E^v_{\text{neg}}$.


Our first objective, the inter-modal contrastive loss, applies a symmetric loss to align visual and textual representations. It associates textual features $E^t$ with paired positive region features $E^v_{\text{pos}}$, while separating them from unpaired samples in the batch:
\vspace{-0.75em}
\begin{equation}
\mathcal{L}_{\,v\, \rightarrow\, t} = -\log \frac{\exp\left(\text{sim}(E^v_{\text{pos}}, E^t) / \tau\right)}{\sum\limits_{\hat{E}^t} \exp\left(\text{sim}(E^v_{\text{pos}}, \hat{E}^t) / \tau\right)}
\end{equation}
\vspace{-0.6em}
\begin{equation}
\mathcal{L}_{\,t\, \rightarrow\, v} = -\log \frac{\exp\left(\text{sim}(E^v_{\text{pos}}, E^t) / \tau\right)}{\sum\limits_{\hat{E}^v_{\text{pos}}} \exp\left(\text{sim}(\hat{E}^v_{\text{pos}}, E^t) / \tau\right)}, \\
\end{equation}
\vspace{-0.6em}
\begin{equation}
\mathcal{L}_{\text{inter}}=\mathcal{L}_{\,v\, \rightarrow\, t} + \mathcal{L}_{\,t\, \rightarrow\, v},
\end{equation}
where $\hat{E}^t$ and $\hat{E}^v_{\text{pos}}$ denote text and visual features from unpaired (in-batch) samples, serving as negative examples for this specific loss.


Our second objective, the intra-image contrastive loss, leverages the fact that only a subset of image regions is relevant. It pushes the instruction $E^t$ closer to the positive region $E^v_{\text{pos}}$ and further from the irrelevant (negative) regions $E^v_{\text{neg}}$ within the same image:
\vspace{-0.75em}
\begin{equation}
\mathcal{L}_{\text{intra}} = -\log \frac{\exp\left(\text{sim}(E^t, E^v_{\text{pos}}) / \tau\right)}{\sum\limits_{E^v_{\text{neg}}} \exp\left(\text{sim}(E^t, E^v_{\text{neg}}) / \tau\right)}.
\end{equation}
The final training objective combines these two losses:
\vspace{-0.75em}
\begin{equation}
\mathcal{L}_{\text{total}} = \mathcal{L}_{\text{inter}} + \lambda  \mathcal{L}_{\text{intra}},
\end{equation}

To optimize this objective, we train only the learnable queries and two MLP layers using these contrastive strategies, while freezing all other components, including the LLM, vision encoder, and projector.

The proposed strategies effectively align the instruction with its corresponding image regions in the common feature space, promoting the selection of the instruction-relevant region through the feature similarity.

\section{PinPoint Dataset}
\label{sec:dataset}

\begin{table*}[t!]
\centering
\renewcommand{\arraystretch}{1.15}
\setlength{\tabcolsep}{3pt} 
\footnotesize
\caption{
Quantitative comparison with state-of-the-art methods on VQA benchmarks: InfoVQA~\cite{infographicvqa}, SPDocVQA~\cite{docvqa}, MPDocVQA~\cite{mpdocvqa}, and GQA~\cite{gqa}. We report Average Normalized Levenshtein Similarity (ANLS) scores, TFLOPs, and latency demonstrating that our proposed method consistently outperforms existing approaches while utilizing a comparable number of visual tokens. For reference, we also include results from a vanilla model that does not apply any token reduction strategy.
}
\label{tab:main}
\resizebox{1.0\linewidth}{!}
{
\begin{tabular}{c|l|cr|cr|cr|cr|c}
\toprule
    \multirow{2}{*}{\textbf{Model}} & \multirow{2}{*}{\textbf{Method}} 
    & \multicolumn{2}{c|}{\textbf{InfoVQA}} 
    & \multicolumn{2}{c|}{\textbf{SPDocVQA}} 
    & \multicolumn{2}{c|}{\textbf{MPDocVQA}}
    & \multicolumn{2}{c|}{\textbf{GQA}} 
    & \multirow{2}{*}{\raisebox{-3.6ex}{\shortstack{\textbf{Latency} \\ \textbf{(ms)}}}}
    \\ \cline{3-10}
    \rule{0pt}{1.25em} & 
    & ANLS $\uparrow$ & FLOPs(T) (\%) $\downarrow$
    & ANLS $\uparrow$ & FLOPs(T) (\%) $\downarrow$
    & ANLS $\uparrow$ & FLOPs(T) (\%) $\downarrow$
    & Acc $\uparrow$ & FLOPs(T) (\%) $\downarrow$ \\
\midrule
    \multirow{6}{*}{\shortstack{LLaVA\\-NeXT-7B}}
    & Vanilla \cite{llavanext} & 0.2552 & 38.98 (100.0\%) & 0.6628 & 51.68 (100.0\%) & 0.3758 & 50.01 (100.0\%) & 0.7598 & 30.39 (100.0\%) & 569.5 \\\cline{2-11}
    \rule{0pt}{1.25em} 
    & Random sampling & 0.2387 & 24.62 (63.2\%) & 0.4888 & 27.70 (53.6\%) & 0.2988 & 26.66 (53.3\%) & 0.7484 & 19.26 (63.4\%) & 360.7 \\
    & ToMe \cite{tome} & 0.1975 & 26.67 (68.4\%) & 0.3215 & 39.41 (76.3\%) & 0.2135 & 38.32 (56.6\%) & 0.7293 & 19.10 (62.9\%) & 315.1 \\
    & FastV \cite{fastv} & 0.2306 & 26.22 (67.3\%) & 0.6099 & 28.10 (54.4\%) & 0.3523 & 29.31 (58.6\%) & 0.7478 & 19.23 (65.9\%) & 386.0 \\
    & PDrop \cite{pyramiddrop} & 0.2335 & 26.00 (66.7\%) & 0.5507 & 26.93 (52.1\%) & 0.3637 & 30.82 (61.6\%) & 0.7436 & 20.33 (66.9\%) & 392.0 \\
    & SparseVLM \cite{sparsevlm} & 0.2428 & 27.45 (70.4\%) & 0.5726 & 32.63 (63.1\%) & 0.3087 & 31.62 (63.2\%) & 0.7449 & 17.85 (58.7\%) & 471.1 \\
    & \cellcolor{gray!10} \textbf{PinPoint (Ours)} & \cellcolor{gray!10} \textbf{0.3024} & \cellcolor{gray!10} 25.48 (65.3\%) & \cellcolor{gray!10} \textbf{0.6472} & \cellcolor{gray!10} 28.44 (55.0\%) & \cellcolor{gray!10} \textbf{0.3866} & \cellcolor{gray!10} 28.28 (56.5\%) & \cellcolor{gray!10} \textbf{0.7608} & \cellcolor{gray!10} 17.96 (59.1\%) & \cellcolor{gray!10} 381.6 \\
\midrule
   \multirow{6}{*}{\shortstack{Qwen2\\-VL-7B}}
    & Vanilla \cite{qwen2vl} & 0.7399 & 51.98 (100\%) & 0.9359 & 73.07 (100\%) & 0.7775 & 55.00 (100.0\%) & 0.7687 & 40.84 (100.0\%) & 907.4 \\\cline{2-11}
    \rule{0pt}{1.25em} 
    & Random sampling & 0.5345 & 29.37 (56.5\%) & 0.8914 & 46.48 (63.6\%) & 0.5651 & 31.04 (56.4\%) & 0.7053 & 20.12 (49.3\%) & 530.4 \\
    & FastV \cite{fastv} & 0.5130 & 31.69 (61.0\%) & 0.8106 & 44.43 (60.8\%) & 0.5561 & 33.51 (60.9\%) & 0.7180 & 19.59 (48.0\%) & 532.7 \\
    & PDrop \cite{pyramiddrop} & 0.4506 & 32.70 (62.9\%) & 0.6848 & 45.85 (62.8\%) & 0.3433 & 29.30 (53.3\%) & 0.5548 & 20.23 (49.5\%) & 500.1 \\
    & \cellcolor{gray!10} \textbf{PinPoint (Ours)} & \cellcolor{gray!10} \textbf{0.7140} & \cellcolor{gray!10} 28.88 (55.5\%) & \cellcolor{gray!10} \textbf{0.8977} & \cellcolor{gray!10} 32.29 (44.2\%) & \cellcolor{gray!10} \textbf{0.6723} & \cellcolor{gray!10} 29.31 (53.3\%) & \cellcolor{gray!10} \textbf{0.7624} & \cellcolor{gray!10} 19.32 (47.3\%) & \cellcolor{gray!10} 555.5 \\
\bottomrule
\end{tabular}
}
\vspace{-1.25em}
\end{table*}

To enable the model to focus on answer-relevant content, it is essential to define the spatial regions associated with correct answers. Although previous studies~\cite{visualcot} have proposed datasets that annotate the bounding box where the answer is located, reasoning over the answer often requires more than just the answer element itself. In many cases, multiple surrounding or supporting elements are necessary to infer the correct answer. Therefore, we provide not only the bounding box for the answer element, but also bounding boxes for the supporting elements required for reasoning, as well as a unified bounding box that encompasses all of them.

Additionally, VLMs still exhibit limited reasoning capabilities on complex images, making their outputs unreliable. As shown in Figure~\ref{fig:dataset} (a), to address this issue, we propose a pipeline for generating high-quality bounding boxes. This pipeline leverages model selection tailored to different data types, enabling the creation of accurate datasets with minimal human effort. All details of the dataset generation pipeline, including the characteristics of the generated datasets and the prompts used, are provided in the supplemental material (see examples in Figure~\ref{fig:dataset} (b)).
\section{Experiments}

\subsection{Datasets} 
To evaluate the effectiveness of our proposed method, we conduct experiments on four challenging Visual Question Answering (VQA) datasets: (i) InfoVQA~\cite{infographicvqa}, which comprises infographic images characterized by high information density, featuring densely packed textual and visual elements. This dataset poses a significant challenge as it requires complex reasoning over fine-grained multimodal information. 
(ii) SPDocVQA~\cite{docvqa}, which includes single-page documents and serves as a benchmark for document-level understanding and question answering.
(iii) MPDocVQA~\cite{mpdocvqa}, which consists of multi-page documents that include diverse content types such as tables, handwritten notes, and diagrams, requiring cross-page reasoning. As LLaVA-NeXT~\cite{llavanext} does not support multi-page inputs, we convert them into a single square-like image by concatenating pages.
(iv) GQA~\cite{gqa}, which is a natural image VQA dataset designed to understand complex questions involving spatial relationships, object attributes, and logical inference. To train on answer-relevant regions, we used the provided ground-truth data for GQA~\cite{gqa}, while the remaining datasets were trained using data labeled according to Sec. \ref{sec:dataset}

\subsection{Implementation and Evaluation Details}
Our method is evaluated on LLaVA-NeXT-Vicuna-7B~\cite{llavanext} and Qwen2-VL-7B~\cite{qwen2vl}. All comparisons utilize the default visual features generated by the respective models. Specifically, for LLaVA-NeXT\cite{llavanext}, only its high-resolution visual features are used. Consistent hyperparameters were used across all datasets. The window size was set to $10 \times 10$, the stride to 7, the target ratio $r$ to 0.6, and $K$ to 100. The models were trained for 5 epochs with a batch size of 32, a learning rate of 2e-5, and $\lambda$ set to 0.5.

To more precisely evaluate the accuracy of our method, the widely used Average Normalized Levenshtein Similarity (ANLS)~\cite{anls} metric is adopted. ANLS measures how similar a predicted answer is to the ground truth by allowing minor differences, such as OCR or spelling errors. In addition, FLOPs (Floating-Point Operations) and average latency in milliseconds are reported to assess computational efficiency, measuring the total computational cost from input processing to the generation of textual outputs. 

Our method is compared against two distinct categories of state-of-the-art approaches: (i) token reduction methods aimed at efficiency, such as ToMe~\cite{tome}, PyramidDrop~\cite{pyramiddrop}, FastV~\cite{fastv}, and SparseVLM~\cite{sparsevlm}, and (ii) region-focusing method aimed at improving accuracy, ViCrop~\cite{mllmsknow}, from which we adopt its rel-attention variant. We also compare against a random token sampling baseline.

\begin{figure*}[t!]
  \centering
  \includegraphics[width=1\linewidth]{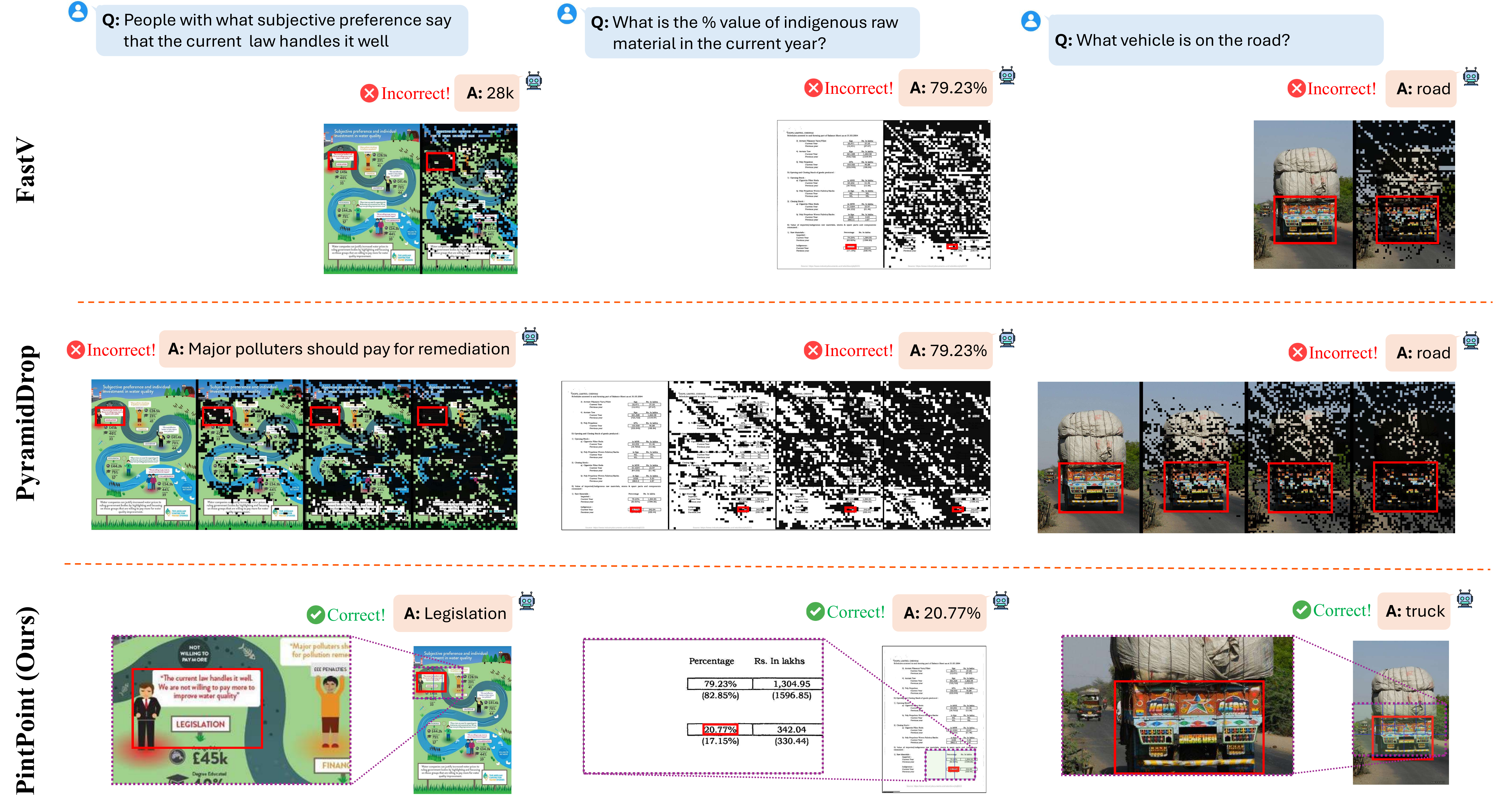}
  \caption{A qualitative comparison among FastV~\cite{fastv}, PyramidDrop~\cite{pyramiddrop}, and our proposed PinPoint reveals distinct behavioral differences. While FastV~\cite{fastv} and PyramidDrop~\cite{pyramiddrop} tend to prune tokens within the answer-relevant regions—often leading to hallucinatory or semantically inconsistent responses—our PinPoint effectively identifies and preserves the answer-critical regions, thereby yielding more accurate and contextually grounded predictions. Note that \textcolor{red}{red} indicates the ground-truth answer regions, \textcolor{blue}{blue} denotes the sliding windows selected by our method, and \textcolor{violet}{purple} highlights the image areas pinpointed as most relevant to the predicted answer.}
  \vspace{-0.5em}
  \label{fig:qualitative}
\end{figure*}
\begin{table*}[t!]
\centering
\renewcommand{\arraystretch}{1.1}
\setlength{\tabcolsep}{6pt}
\footnotesize
\caption{
\textbf{Performance Analysis of Focused Region Selection.} PinPoint effectively selects and provides answer-relevant regions to the LLM, resulting in higher accuracy with fewer tokens compared to the vanilla baseline~\cite{llavanext}. Conversely, ViCrop~\cite{mllmsknow} introduces irrelevant regions that act as noise, causing performance degradation despite a high computational cost.}
\label{tab:main_con}
\resizebox{1.0\linewidth}{!}
{
\begin{tabular}{l|cc|cc|cc|cc}
\toprule
\multirow{2}{*}{\textbf{\shortstack{Model}}} 
& \multicolumn{2}{c|}{\textbf{InfoVQA}} 
& \multicolumn{2}{c|}{\textbf{SPDocVQA}} 
& \multicolumn{2}{c|}{\textbf{MPDocVQA}} 
& \multicolumn{2}{c}{\textbf{GQA}} \\
\cline{2-9}
\rule{0pt}{1.25em} 
& ANLS $\uparrow$ & FLOPs(T) $\downarrow$ & ANLS $\uparrow$ & FLOPs(T) $\downarrow$ & ANLS $\uparrow$ & FLOPs(T) $\downarrow$ & ANLS $\uparrow$ & FLOPs(T) $\downarrow$ \\
\midrule
Vanilla~\cite{llavanext} & 0.2552 & 38.98 (100.0\%) & 0.6628 & 51.68 (100.0\%) & 0.3758 & 50.01 (100.0\%) & 0.7598 & 30.39 (100\%) \\
ViCrop~\cite{mllmsknow} & 0.2547 & 147.39 (378.1\%) & 0.5941 & 168.71 (168.7\%) & 0.3661 & 171.86 (343.6\%) & 0.7497 & 137.58 (452.71\%) \\
\rowcolor{gray!10} \textbf{Ours} & 0.3024 & 25.48 (65.3\%) & 0.6472 & 28.44 (55.0\%) & 0.3866 & 28.28 (56.5\%) & 0.7608 & 17.96 (59.1\%) \\
\rowcolor{gray!10} \textbf{Ours} $+$ \textbf{Global} & \textbf{0.3075} & 33.09 (84.9\%) & \textbf{0.6732} & 37.34 (72.3\%) & \textbf{0.3976} & 37.20 (74.4\%) & \textbf{0.7704} & 26.64 (87.7\%) \\
\bottomrule
\end{tabular}
}
\vspace{-1.75em}
\end{table*}

\subsection{Quantitative Analysis}
\paragraph{Comparison to State-of-the-art Approaches.}
As illustrated in Table~\ref{tab:main}, our model, PinPoint when applied to LLaVA-NeXT~\cite{llavanext}, achieves the best performance, consistently outperforming all competing methods. Moreover, on InfoVQA~\cite{llavanext}, MPDocVQA~\cite{mpdocvqa}, and GQA~\cite{gqa}, PinPoint even surpasses the vanilla model~\cite{llavanext} that processes all visual tokens. On InfoVQA~\cite{infographicvqa}, this corresponds to 18.5\% higher accuracy with only 65.3\% of the computational cost.
On the same benchmark, PinPoint also outperforms SparseVLM~\cite{sparsevlm}, the strongest prior method, by 24.5\% accuracy while using 4.9\% fewer FLOPs.
Notably, this performance gap widens on more challenging, high-resolution, and information-dense images.

Integrating PinPoint with Qwen2-VL~\cite{qwen2vl} further validates our approach, yielding the highest performance among all competing methods. In contrast, FastV~\cite{fastv} and PyramidDrop~\cite{pyramiddrop} show substantial performance degradation, indicating that retaining tokens based on intermediate layer attention weights is a flawed strategy that can lead to a severe drop in accuracy.

\begin{table*}[t]
\caption{
\textbf{Effect of the intra-image contrastive loss ($\mathcal{L}_{intra}$).}
Including the $\mathcal{L}_{intra}$ loss component ($\checkmark$) consistently improves both ANLS and Region Accuracy scores compared to the baseline (--) across all models and datasets.
}
\label{tab:ablation_loss}
\vspace{-0.5em}
\centering
\renewcommand{\arraystretch}{1}
\setlength{\tabcolsep}{5pt}
\footnotesize
{
\begin{tabular}{ c | c | cc | cc | cc | cc}
\toprule
\multirow{2}{*}{\textbf{Model}} &
\multirow{2}{*}{$\mathcal{L}_{intra}$} &
\multicolumn{2}{c|}{\textbf{InfoVQA}} &
\multicolumn{2}{c|}{\textbf{SPDocVQA}} &
\multicolumn{2}{c|}{\textbf{MPDocVQA}} &
\multicolumn{2}{c}{\textbf{GQA}} \\
\cline{3-10}
\rule{0pt}{1.25em} & &
ANLS$\uparrow$ & Region Acc.$\uparrow$ &
ANLS$\uparrow$ & Region Acc.$\uparrow$ &
ANLS$\uparrow$ & Region Acc.$\uparrow$ &
ANLS$\uparrow$ & Region Acc.$\uparrow$ \\
\midrule
\multirow{2}{*}{\shortstack{LLaVA\\-NeXT-7B}} 
& -- & 0.3011 & 82\% & 0.6465 & 96\% & 0.3811 & 85\% & 0.7465 & 93\% \\
& \checkmark & 0.3024 & 84\% & 0.6472 & 98\% & 0.3866 & 87\% & 0.7608 & 98\% \\
\midrule
\multirow{2}{*}{\shortstack{Qwen2\\-VL-7B}} 
& -- & 0.7010 & 92\% & 0.8799 & 96\% & 0.6595 & 90\% & 0.7600 & 97\% \\
& \checkmark & 0.7140 & 95\% & 0.8977 & 98\% & 0.6723 & 94\% & 0.7624 & 98\% \\
\bottomrule
\end{tabular}
}
\vspace{-0.5em}
\end{table*}



\begin{table*}[t!]
\caption{
\textbf{Effect of Region Refinement Module.} 
By capturing more fine-grained and discriminative visual cues, Region Refinement allows the model to significantly outperform the baseline. 
}
\label{tab:region_refine}
\vspace{-0.5em}
\centering
\renewcommand{\arraystretch}{1}
\setlength{\tabcolsep}{7pt}
\footnotesize
{
\begin{tabular}{c|cc|cc|cc|cc}
\toprule
\multirow{2}{*}{\textbf{\shortstack{Region \\ Refinement}}} 
& \multicolumn{2}{c|}{\textbf{InfoVQA}} 
& \multicolumn{2}{c|}{\textbf{SPDocVQA}} 
& \multicolumn{2}{c|}{\textbf{MPDocVQA}} 
& \multicolumn{2}{c}{\textbf{GQA}} \\
\cline{2-9} 
\rule{0pt}{1.25em} 
& ANLS & Region Acc. & ANLS & Region Acc. & ANLS & Region Acc. & ANLS & Region Acc. \\ 
\midrule
-- & 0.2603 & \multirow{2}{*}{84\%} & 0.6230 & \multirow{2}{*}{98\%} & 0.3403 & \multirow{2}{*}{87\%} & 0.7484 & \multirow{2}{*}{98\%} \\
\checkmark & \textbf{0.3024} &  & \textbf{0.6472} &  & \textbf{0.3866} &  & \textbf{0.7608} &  \\ 
\bottomrule
\end{tabular}
}
\vspace{-1.5em}
\end{table*}

\paragraph{Comparison of Region-Focusing Methods.} Table~\ref{tab:main_con} presents the performance results when focusing on query-relevant regions using LLaVA-NeXT~\cite{llavanext}. In this analysis, "Ours + Global" refers to a setting where the LLM receives both the re-encoded tokens from the refined regions (as in our standard approach) and the initial low-resolution visual tokens from the first encoding pass, which represent the global image context. 

Our method achieves superior performance compared to ViCrop \cite{mllmsknow}. Moreover, "Ours + Global" consistently attains the highest performance across all datasets while maintaining substantial computational efficiency relative to the vanilla model~\cite{llavanext}. ViCrop \cite{mllmsknow}, in contrast, not only shows noticeable performance degradation compared to the vanilla baseline~\cite{llavanext} but also incurs considerable computational overhead due to its multiple VLM passes to identify relevant regions and process additional tokens. These findings indicate that our model more effectively localizes instruction-relevant regions and produces accurate answers, while remaining computationally efficient.


\subsection{Qualitative Analysis}
We provide a qualitative comparison among FastV \cite{fastv}, PyramidDrop \cite{pyramiddrop}, and our proposed PinPoint in Figure \ref{fig:qualitative} for intuitive understanding. PyramidDrop~\cite{pyramiddrop} and FastV~\cite{fastv} frequently prune tokens within the answer-relevant regions—often resulting in hallucinatory or semantically inconsistent responses. In contrast, our method effectively localizes instruction-relevant regions. These results confirm that PinPoint successfully selects regions that are semantically aligned with the instruction, thereby enabling the model to generate accurate answers aligned with visual evidence. 

\vspace{-0.3em}
\subsection{Efficiency of PinPoint in Training}
PinPoint does not involve fine-tuning or backpropagation through the LLM. Instead, it operates solely on the visual features extracted from the frozen vision encoder and projector by applying a lightweight MLP and attention operations with a set of learnable guidance query tokens. This design enables highly efficient training while achieving performance that is competitive with, or even superior to, baseline models. 

We detail the training configuration for LLaVA-NeXT~\cite{llavanext}. As shown in Table~\ref{tab:param_comparison}, the total number of trainable parameters accounts for 1.4\% of the entire model. When training on the InfoVQA~\cite{infographicvqa} dataset, the model converges in 3 epochs with a batch size of 32, requiring a total of 1.33 GPU hours for the entire training run. All experiments were performed on two NVIDIA A100 GPUs. 


\vspace{-0.3em}
\subsection{Ablation Studies}
\paragraph{Effectiveness of Intra-Image Contrastive Loss.}

To evaluate the impact of the intra-image contrastive loss ($\mathcal{L}_{intra}$), we conduct an ablation study comparing variants of our model with and without this loss term across VQA benchmarks. As shown in Table~\ref{tab:ablation_loss}, the inclusion of $\mathcal{L}_{intra}$ contributes significantly to performance, yielding the best results in accurately identifying instruction-relevant image regions and thereby enabling accurate answer generation.

\begin{table}[t!]
\caption{
\textbf{Parameter Efficiency of PinPoint.} The table details the number of trainable parameters (our PinPoint module) relative to the total parameters of the frozen base models. In both cases, the trainable parameters constitute less than 1.5\% of the entire model, highlighting the high efficiency of our method.
}
\label{tab:param_comparison}
\centering
\renewcommand{\arraystretch}{1}
\setlength{\tabcolsep}{7pt} 
\footnotesize
{
\vspace{-0.5em}
\begin{tabular}{l|c|c}
\toprule
\textbf{Model} & \textbf{Trainable Params} & \textbf{Percent of Total} \\
\midrule
LLaVA-NeXT-7B~\cite{llavanext} & 99.4 M & 1.40\% \\
Qwen2-VL-7B~\cite{qwen2vl}   & 72.6 M  & 0.86\% \\ 
\bottomrule
\end{tabular}
}
\vspace{-1.75em}
\end{table}

\paragraph{Effectiveness of Region Refinement.}

To evaluate the effectiveness of the Region Refinement stage, we conduct an ablation study on LLaVA-NeXT~\cite{llavanext}, comparing our method with a variant that uses only the initially encoded (and potentially contaminated) region tokens. As shown in Table~\ref{tab:region_refine}, the model with Region Refinement substantially outperforms the variant relying solely on the initial visual tokens extracted from the first vision-encoder pass. This demonstrates that Region Refinement enables the model to recover more fine-grained and discriminative visual cues, ultimately leading to improved answer accuracy.
\vspace{-0.3em}
\section{Conclusion}
This paper proposes PinPoint, a novel model for instruction-relevant visual token selection and refinement. This approach first identifies instruction-relevant image regions and then refines these regions to extract fine-grained visual tokens. To evaluate the effectiveness of our model, we construct a new dataset built upon existing VQA benchmarks annotated with instruction-relevant regions likely to contain answers or supporting evidence. Using this dataset, PinPoint is compared against state-of-the-art methods on challenging VQA benchmarks, demonstrating superior performance across all tasks.
\section*{Acknowledgement}
This work was the result of project supported by KT(Korea Telecom) - Korea University AICT R\&D Cener. This work was partly supported by Institute of Information \& communications Technology Planning \& Evaluation(IITP) under the Leading Generative AI Human Resources Development(IITP-2025-RS-2024-00397085, 10\%) grant, the artificial intelligence star fellowship support program to nurture the best talents (IITP-2025-RS-2025-02304828, 30\%) grant, and ITRC(Information Technology Research Center) grant (IITP-2025-RS-2022-00156295, 10\%) funded by the Korea government(MSIT).
{
    \small
    \bibliographystyle{ieeenat_fullname}
    \bibliography{main}

@String(CVPR= {IEEE Conf. Comput. Vis. Pattern Recog.})

@String(ICCV= {Int. Conf. Comput. Vis.})

@String(ECCV= {Eur. Conf. Comput. Vis.})

@String(ICLR = {Int. Conf. Learn. Represent.})

@String(AAAI = {AAAI})

@String(CVPR  = {CVPR})

@String(ICCV  = {ICCV})

@String(ECCV  = {ECCV})

@String(ICLR  = {ICLR})

@inproceedings{pyramiddrop,
  title={Pyramiddrop: Accelerating your large vision-language models via pyramid visual redundancy reduction},
  author={Xing, Long and Huang, Qidong and Dong, Xiaoyi and Lu, Jiajie and Zhang, Pan and Zang, Yuhang and Cao, Yuhang and He, Conghui and Wang, Jiaqi and Wu, Feng and others},
  booktitle={CVPR},
  year={2025}
}

@inproceedings{infographicvqa,
  title={Infographicvqa},
  author={Mathew, Minesh and Bagal, Viraj and Tito, Rub{\`e}n and Karatzas, Dimosthenis and Valveny, Ernest and Jawahar, CV},
  booktitle={Proceedings of the IEEE/CVF Winter Conference on Applications of Computer Vision},
  year={2022}
}

@article{llavanext,
    title={Llava-next-interleave: Tackling multi-image, video, and 3d in large multimodal models},
    author={Li, Feng and Zhang, Renrui and Zhang, Hao and Zhang, Yuanhan and Li, Bo and Li, Wei and Ma, Zejun and Li, Chunyuan},
    journal={arXiv preprint arXiv:2407.07895},
    year={2024}
}

@inproceedings{fastv,
  title={An image is worth 1/2 tokens after layer 2: Plug-and-play inference acceleration for large vision-language models},
  author={Chen, Liang and Zhao, Haozhe and Liu, Tianyu and Bai, Shuai and Lin, Junyang and Zhou, Chang and Chang, Baobao},
  booktitle={ECCV},
  year={2024}
}

@inproceedings{tome,
  title={Token Merging: Your ViT But Faster},
  author={Bolya, Daniel and Fu, Cheng-Yang and Dai, Xiaoliang and Zhang, Peizhao and Feichtenhofer, Christoph and Hoffman, Judy},
  booktitle={ICLR},
  year={2023}
}

@inproceedings{blip2,
  title={BLIP-2: bootstrapping language-image pre-training with frozen image encoders and large language models},
  author={Li, Junnan and Li, Dongxu and Savarese, Silvio and Hoi, Steven},
  booktitle={ICML},
  year={2023}
}

@article{madtp,
  title={MADTP: Multimodal Alignment-Guided Dynamic Token Pruning for Accelerating Vision-Language Transformer},
  author={Jianjian, Cao and Peng, Ye and Shengze, Li and Chong, Yu and Yansong, Tang and Jiwen, Lu and Tao, Chen},
  journal={CVPR},
  year={2024}
}

@inproceedings{ivtp,
  title={IVTP: Instruction-Guided Visual Token Pruning for Large Vision-Language Models},
  author={Huang, Kai and Zou, Hao and Xi, Ye and Wang, BoChen and Xie, Zhen and Yu, Liang},
  booktitle={ECCV},
  year={2024}
}

@inproceedings{bpe,
  title={Neural machine translation of rare words with subword units},
  author={Sennrich, Rico and Haddow, Barry and Birch, Alexandra},
  booktitle={ACL},
  year={2016}
}

@inproceedings{vit,
  title={An image is worth 16x16 words: Transformers for image recognition at scale},
  author={Dosovitskiy, Alexey and Beyer, Lucas and Kolesnikov, Alexander and Weissenborn, Dirk and Zhai, Xiaohua and Unterthiner, Thomas and Dehghani, Mostafa and Minderer, Matthias and Heigold, Georg and Gelly, Sylvain and others},
  booktitle={ICLR},
  year={2021}
}

@article{qwen,
  title={Qwen technical report},
  author={Bai, Jinze and Bai, Shuai and Chu, Yunfei and Cui, Zeyu and Dang, Kai and Deng, Xiaodong and Fan, Yang and Ge, Wenbin and Han, Yu and Huang, Fei and others},
  journal={arXiv preprint arXiv:2309.16609},
  year={2023}
}

@inproceedings{improvedllava,
  title={Improved baselines with visual instruction tuning},
  author={Liu, Haotian and Li, Chunyuan and Li, Yuheng and Lee, Yong Jae},
  booktitle={CVPR},
  year={2024}
}

@inproceedings{videollama,
  title={Video-LLaMA: An Instruction-tuned Audio-Visual Language Model for Video Understanding},
  author={Zhang, Hang and Li, Xin and Bing, Lidong},
  booktitle={EMNLP (Demos)},
  year={2023}
}

@inproceedings{minigpt,
  title={MiniGPT-4: Enhancing Vision-Language Understanding with Advanced Large Language Models},
  author={Zhu, Deyao and Chen, Jun and Shen, Xiaoqian and Li, Xiang and Elhoseiny, Mohamed},
  booktitle={ICLR},
  year={2024}
}

@inproceedings{videochatgpt,
    title={Video-ChatGPT: Towards Detailed Video Understanding via Large Vision and Language Models},
    author={Maaz, Muhammad and Rasheed, Hanoona and Khan, Salman and Khan, Fahad Shahbaz},
    booktitle={ACL},
    year={2024}
}

@article{textmonkey,
  title={TextMonkey: An OCR-Free Large Multimodal Model for Understanding Document},
  author={Liu, Yuliang and Yang, Biao and Liu, Qiang and Li, Zhang and Ma, Zhiyin and Zhang, Shuo and Bai, Xiang},
  journal={CoRR},
  year={2024}
}

@inproceedings{mplugdocowl15,
  title={mPLUG-DocOwl 1.5: Unified Structure Learning for OCR-free Document Understanding},
  author={Hu, Anwen and Xu, Haiyang and Ye, Jiabo and Yan, Ming and Zhang, Liang and Zhang, Bo and Zhang, Ji and Jin, Qin and Huang, Fei and Zhou, Jingren},
  booktitle={EMNLP(Findings)},
  year={2024}
}

@inproceedings{docvqa,
  title={DocVQA: A Dataset for VQA on Document Images},
  author={Mathew, Minesh and Karatzas, Dimosthenis and Jawahar, CV},
  booktitle={IEEE},
  year={2021}
}

@article{mpdocvqa,
  title={Hierarchical multimodal transformers for multipage docvqa},
  author={Tito, Rub{\`e}n and Karatzas, Dimosthenis and Valveny, Ernest},
  journal={Pattern Recognition},
  year={2023}
}

@inproceedings{anls,
  title={Scene text visual question answering},
  author={Biten, Ali Furkan and Tito, Ruben and Mafla, Andres and Gomez, Lluis and Rusinol, Mar{\c{c}}al and Valveny, Ernest and Jawahar, CV and Karatzas, Dimosthenis},
  booktitle={ICCV},
  year={2019}
}

@inproceedings{visualcot,
      title={Visual CoT: Advancing Multi-Modal Language Models with a Comprehensive Dataset and Benchmark for Chain-of-Thought Reasoning}, 
      author={Hao Shao and Shengju Qian and Han Xiao and Guanglu Song and Zhuofan Zong and Letian Wang and Yu Liu and Hongsheng Li},
      booktitle={NeurIPS},
      year={2024}
}

@inproceedings{monkey,
  title={Monkey: Image Resolution and Text Label are Important Things for Large Multi-Modal Models},
  author={Li, Zhang and Yang, Biao and Liu, Qiang and Ma, Zhiyin and Zhang, Shuo and Yang, Jingxu and Sun, Yabo and Liu, Yuliang and Bai, Xiang},
  booktitle={CVPR},
  year={2024}
}

@article{internvl15,
  title={How far are we to gpt-4v? closing the gap to commercial multimodal models with open-source suites},
  author={Chen, Zhe and Wang, Weiyun and Tian, Hao and Ye, Shenglong and Gao, Zhangwei and Cui, Erfei and Tong, Wenwen and Hu, Kongzhi and Luo, Jiapeng and Ma, Zheng and others},
  journal={ Sci. China Inf. Sci.},
  year={2024},
}

@inproceedings{chen2024internvlscalingvisionfoundation,
    title={Internvl: Scaling up vision foundation models and aligning for generic visual-linguistic tasks},
    author={Chen, Zhe and Wu, Jiannan and Wang, Wenhai and Su, Weijie and Chen, Guo and Xing, Sen and Zhong, Muyan and Zhang, Qinglong and Zhu, Xizhou and Lu, Lewei and others},
    booktitle={CVPR},
    year={2024}
}

@inproceedings{fitprune,
  title={Fit and prune: Fast and training-free visual token pruning for multi-modal large language models},
  author={Ye, Weihao and Wu, Qiong and Lin, Wenhao and Zhou, Yiyi},
  booktitle={AAAI},
  year={2025}
}

@article{qwen2vl,
  title={Qwen2-vl: Enhancing vision-language model's perception of the world at any resolution},
  author={Wang, Peng and Bai, Shuai and Tan, Sinan and Wang, Shijie and Fan, Zhihao and Bai, Jinze and Chen, Keqin and Liu, Xuejing and Wang, Jialin and Ge, Wenbin and others},
  journal={arXiv preprint arXiv:2409.12191},
  year={2024}
}

@article{qwen25vl,
  title={Qwen2. 5-vl technical report},
  author={Bai, Shuai and Chen, Keqin and Liu, Xuejing and Wang, Jialin and Ge, Wenbin and Song, Sibo and Dang, Kai and Wang, Peng and Wang, Shijie and Tang, Jun and others},
  journal={arXiv preprint arXiv:2502.13923},
  year={2025}
}

@article{gpt4,
  title={Gpt-4 technical report},
  author={Achiam, Josh and Adler, Steven and Agarwal, Sandhini and Ahmad, Lama and Akkaya, Ilge and Aleman, Florencia Leoni and Almeida, Diogo and Altenschmidt, Janko and Altman, Sam and Anadkat, Shyamal and others},
  journal={arXiv preprint arXiv:2303.08774},
  year={2023}
}

@inproceedings{sparsevlm,
  title={SparseVLM: Visual Token Sparsification for Efficient Vision-Language Model Inference},
  author={Zhang, Yuan and Fan, Chun-Kai and Ma, Junpeng and Zheng, Wenzhao and Huang, Tao and Cheng, Kuan and Gudovskiy, Denis and Okuno, Tomoyuki and Nakata, Yohei and Keutzer, Kurt and others},
  booktitle={ICML},
  year={2025}
}

@inproceedings{pai,
  title={Paying more attention to image: A training-free method for alleviating hallucination in lvlms},
  author={Liu, Shi and Zheng, Kecheng and Chen, Wei},
  booktitle={ECCV},
  pages={125--140},
  year={2024},
  organization={Springer}
}

@inproceedings{vissink,
  title={See what you are told: Visual attention sink in large multimodal models},
  author={Kang, Seil and Kim, Jinyeong and Kim, Junhyeok and Hwang, Seong Jae},
  booktitle={ICLR},
  year={2025}
}

@inproceedings{sparc,
  title={Visual Attention Never Fades: Selective Progressive Attention ReCalibration for Detailed Image Captioning in Multimodal Large Language Models},
  author={Jung, Mingi and Lee, Saehyung and Kim, Eunji and Yoon, Sungroh},
  booktitle={ICML},
  year={2025}
}

@inproceedings{instructblip,
  title={InstructBLIP: towards general-purpose vision-language models with instruction tuning},
  author={Dai, Wenliang and Li, Junnan and Li, Dongxu and Tiong, Anthony Meng Huat and Zhao, Junqi and Wang, Weisheng and Li, Boyang and Fung, Pascale and Hoi, Steven},
  booktitle={NeurIPS},
  pages={49250--49267},
  year={2023}
}

@inproceedings{llava-prumerge,
title={LLaVA-PruMerge: Adaptive Token Reduction for Efficient Large Multimodal Models},
author={Yuzhang Shang and Mu Cai and Bingxin Xu and Yong Jae Lee and Yan Yan},
booktitle={ICCV},
year={2025}
}

@inproceedings{seal,
  title={V*: Guided Visual Search as a Core Mechanism in Multimodal LLMs},
  author={Wu, Penghao and Xie, Saining},
  booktitle={CVPR},
  pages={13084--13094},
  year={2024},
  organization={IEEE}
}

@inproceedings{mllmsknow,
  title={Mllms know where to look: Training-free perception of small visual details with multimodal llms},
  author={Zhang, Jiarui and Khayatkhoei, Mahyar and Chhikara, Prateek and Ilievski, Filip},
  booktitle={ICLR},
  year={2025}
}

@book{whereswally,
  author       = {Handford, Martin},
  illustrator  = {Handford, Martin},
  title        = {Where's Wally?},
  year         = {1987},
  language     = {English},
  subject      = {Where's Wally?},
  publisher    = {Walker Books},
  address      = {United Kingdom},
  note         = {US editions published by Little, Brown and later Candlewick Press. Media type: Print (hardback).},
  pages        = {26},
  isbn         = {0-316-34293-9},
  oclc         = {15109312},
  lccn         = {PZ7.H1918 Wh 1987},
  month        = {June},
  day          = {25},
}

@inproceedings{gqa,
  title={Gqa: A new dataset for real-world visual reasoning and compositional question answering},
  author={Hudson, Drew A and Manning, Christopher D},
  booktitle={CVPR},
  pages={6700--6709},
  year={2019}
}

@inproceedings{textvqa,
  title={Towards vqa models that can read},
  author={Singh, Amanpreet and Natarajan, Vivek and Shah, Meet and Jiang, Yu and Chen, Xinlei and Batra, Dhruv and Parikh, Devi and Rohrbach, Marcus},
  booktitle={CVPR},
  year={2019}
}

@inproceedings{vqa,
  title={Making the V in VQA Matter: Elevating the Role of Image Understanding in Visual Question Answering},
  author={Goyal, Yash and Khot, Tejas and Summers-Stay, Douglas and Batra, Dhruv and Parikh, Devi},
  booktitle={CVPR},
  pages={6325--6334},
  year={2017},
  organization={IEEE Computer Society}
}

@inproceedings{chartqa,
  title={Chartqa: A benchmark for question answering about charts with visual and logical reasoning},
  author={Masry, Ahmed and Do, Xuan Long and Tan, Jia Qing and Joty, Shafiq and Hoque, Enamul},
  booktitle={Findings of the association for computational linguistics: ACL 2022},
  pages={2263--2279},
  year={2022}
}

@article{pope,
  title={Evaluating object hallucination in large vision-language models},
  author={Li, Yifan and Du, Yifan and Zhou, Kun and Wang, Jinpeng and Zhao, Wayne Xin and Wen, Ji-Rong},
  journal={arXiv preprint arXiv:2305.10355},
  year={2023}
}

@article{chair,
  title={Object hallucination in image captioning},
  author={Rohrbach, Anna and Hendricks, Lisa Anne and Burns, Kaylee and Darrell, Trevor and Saenko, Kate},
  journal={arXiv preprint arXiv:1809.02156},
  year={2018}
}

@inproceedings{mscoco,
  title={Microsoft coco: Common objects in context},
  author={Lin, Tsung-Yi and Maire, Michael and Belongie, Serge and Hays, James and Perona, Pietro and Ramanan, Deva and Doll{\'a}r, Piotr and Zitnick, C Lawrence},
  booktitle={European conference on computer vision},
  pages={740--755},
  year={2014},
  organization={Springer}
}

@article{amber,
  title={Amber: An llm-free multi-dimensional benchmark for mllms hallucination evaluation},
  author={Wang, Junyang and Wang, Yuhang and Xu, Guohai and Zhang, Jing and Gu, Yukai and Jia, Haitao and Wang, Jiaqi and Xu, Haiyang and Yan, Ming and Zhang, Ji and others},
  journal={arXiv preprint arXiv:2311.07397},
  year={2023}
}

@inproceedings{honeybee,
  title={Honeybee: Locality-enhanced projector for multimodal llm},
  author={Cha, Junbum and Kang, Wooyoung and Mun, Jonghwan and Roh, Byungseok},
  booktitle={CVPR},
  pages={13817--13827},
  year={2024}
}

@inproceedings{RegionVLM,
  title={Toward interactive regional understanding in vision-large language models},
  author={Lee, Jungbeom and Chun, Sanghyuk and Yun, Sangdoo},
  booktitle={NAACL},
  pages={6416--6429},
  year={2024}
}

@inproceedings{weaklyseg,
  title={Weakly supervised referring image segmentation with intra-chunk and inter-chunk consistency},
  author={Lee, Jungbeom and Lee, Sungjin and Nam, Jinseok and Yu, Seunghak and Do, Jaeyoung and Taghavi, Tara},
  booktitle={ICCV},
  pages={21870--21881},
  year={2023}
}

@inproceedings{mmmu,
  title={Mmmu: A massive multi-discipline multimodal understanding and reasoning benchmark for expert agi},
  author={Yue, Xiang and Ni, Yuansheng and Zhang, Kai and Zheng, Tianyu and Liu, Ruoqi and Zhang, Ge and Stevens, Samuel and Jiang, Dongfu and Ren, Weiming and Sun, Yuxuan and others},
  booktitle={CVPR},
  pages={9556--9567},
  year={2024}
}

@inproceedings{mmmu-pro,
  title={Mmmu-pro: A more robust multi-discipline multimodal understanding benchmark},
  author={Yue, Xiang and Zheng, Tianyu and Ni, Yuansheng and Wang, Yubo and Zhang, Kai and Tong, Shengbang and Sun, Yuxuan and Yu, Botao and Zhang, Ge and Sun, Huan and others},
  booktitle={ACL},
  pages={15134--15186},
  year={2025}
}
}
\clearpage

\renewcommand{\thefigure}{A\arabic{figure}}
\setcounter{figure}{0}
\renewcommand{\thetable}{A\arabic{table}}
\setcounter{table}{0}
\renewcommand{\thesection}{\Alph{section}}
\setcounter{section}{0}

\maketitlesupplementary

\section{Experimental Details}
\label{apdx:apdx_experiment}

\subsection{Implementation Details}
\label{apdx:apdx_implementation}
\myparagraph{LLaVA-NeXT.} LLaVA-NeXT~\cite{llavanext} employs the Dynamic High Resolution (AnyRes) module, which processes high-resolution images by dividing them into multiple patches and encoding them independently. In all experiments, we follow the same configuration and allow AnyRes to generate at most 6 patches per image. During the Region Selection stage, approximately 60\% of the original image area is selected and re-encoded. Since Region Refinement operates on this smaller cropped area, we restrict AnyRes to at most 4 patches in this stage, which is sufficient to capture finer-grained visual details while further reducing computational cost.

\myparagraph{Qwen2-VL.}
The Qwen2-VL~\cite{qwen2vl} model processes visual tokens by adapting to the scale of the input image using its Native Dynamic Resolution method. As in the LLaVA-NeXT experiments, the region refinement stage only re-processes a cropped subset of the image. Therefore, during this re-encoding step, we restrict the vision processor’s maximum processed pixels to 60\% of the original setting, which effectively reduces the computational cost while still preserving sufficient resolution to capture fine-grained visual details.

\section{Additional Experiments}
\subsection{Efficiency Analysis of Instruction-Region Alignment}
We analyze the computational overhead introduced by the PinPoint module, which performs region selection and region refinement using a small set of learnable queries. Table~\ref{tab:flops_comparison} reports the FLOPs of PinPoint itself and the average total FLOPs across all VQA datasets~\cite{infographicvqa, docvqa, mpdocvqa, gqa} when PinPoint is adopted. For LLaVA-NeXT~\cite{llavanext}, PinPoint adds only 1.67T FLOPs, corresponding to 7.14\% of the total 23.37T FLOPs. For Qwen2-VL~\cite{qwen2vl}, the overhead is 5.68T FLOPs (22.3\% of 25.44T), reflecting that its Native Dynamic Resolution scheme feeds more visual tokens into the ViT~\cite{vit} backbone during region refinement than LLaVA-NeXT~\cite{llavanext}. This result confirms that PinPoint effectively locates instruction-relevant regions with minimal additional computation, maintaining a significantly lower operational cost compared to vanilla model.







\begin{table}[t!]
\caption{
\textbf{Computational Overhead of the PinPoint Module.}
The table reports FLOPs of the frozen model (Vanilla), FLOPs (T) with PinPoint, and the resulting ratio.
}
\label{tab:flops_comparison}
\centering
\renewcommand{\arraystretch}{1}
\setlength{\tabcolsep}{6.5pt}
\footnotesize
{
\begin{tabular}{
    l|c|c|c|c
}
\toprule

\multirow{2}{*}{\textbf{\rule{0pt}{1.2em} Model}}
& \textbf{Vanilla}
& \multicolumn{3}{c}{\textbf{Ours}} \\

\cline{2-5}

\rule{0pt}{1.25em} & \textbf{Total} 
& \textbf{PinPoint} 
& \textbf{Total} 
& \textbf{Ratio} \\

\midrule

LLaVA-NeXT-7B & 40.30 & 1.67 & 23.37 & 7.14\% \\
Qwen2-VL-7B   & 52.30 & 5.68 & 25.44 & 22.30\% \\

\bottomrule
\end{tabular}
}
\end{table}

\begin{table}[t!]
\centering
\renewcommand{\arraystretch}{1.15}
\setlength{\tabcolsep}{4pt}
\footnotesize
\caption{
\textbf{Performance Analysis of Focused Region Selection for Qwen2-VL~\cite{qwen2vl}.} 
PinPoint effectively identifies and supplies answer-relevant regions, thereby improving performance on VQA tasks. Metric: ANLS. \(r\): 0.3.}
\label{tab:qwen_con}
{
\begin{tabular}{l|c|c|c|c}
\toprule
\textbf{Model} 
& \textbf{InfoVQA} 
& \textbf{SPDocVQA} 
& \textbf{MPDocVQA} 
& \textbf{GQA} \\
\midrule

Vanilla~\cite{qwen2vl} 
& 0.7399
& 0.9359
& 0.7775
& \textbf{0.7687} \\


\rowcolor{gray!10} \textbf{Ours + Global}
& \textbf{0.7560}
& \textbf{0.9380}
& \textbf{0.8071}
& 0.7682 \\
\bottomrule
\end{tabular}
}
\vspace{-1.5em}
\end{table}

\subsection{Quantitative Results}
\myparagraph{Effectiveness of Providing Pinpointed Regions to Qwen2-VL.} Table~\ref{tab:qwen_con} reports the results of augmenting Qwen2-VL~\cite{qwen2vl} with both the full image (Global) and the answer-related crops identified by PinPoint. Unlike our main experiments, where only the PinPoint-selected regions are provided, here the model receives the global image together with the localized instruction-relevant regions. The results show substantial performance gains on fine-grained VQA datasets that require attending to small visual details (e.g., InfoVQA, MPDocVQA), confirming that PinPoint effectively localizes answer-related regions and that explicitly highlighting these regions enables Qwen2-VL to generate more accurate answers.

\begin{figure*}[t!]
  \centering
  \includegraphics[width=1\linewidth]{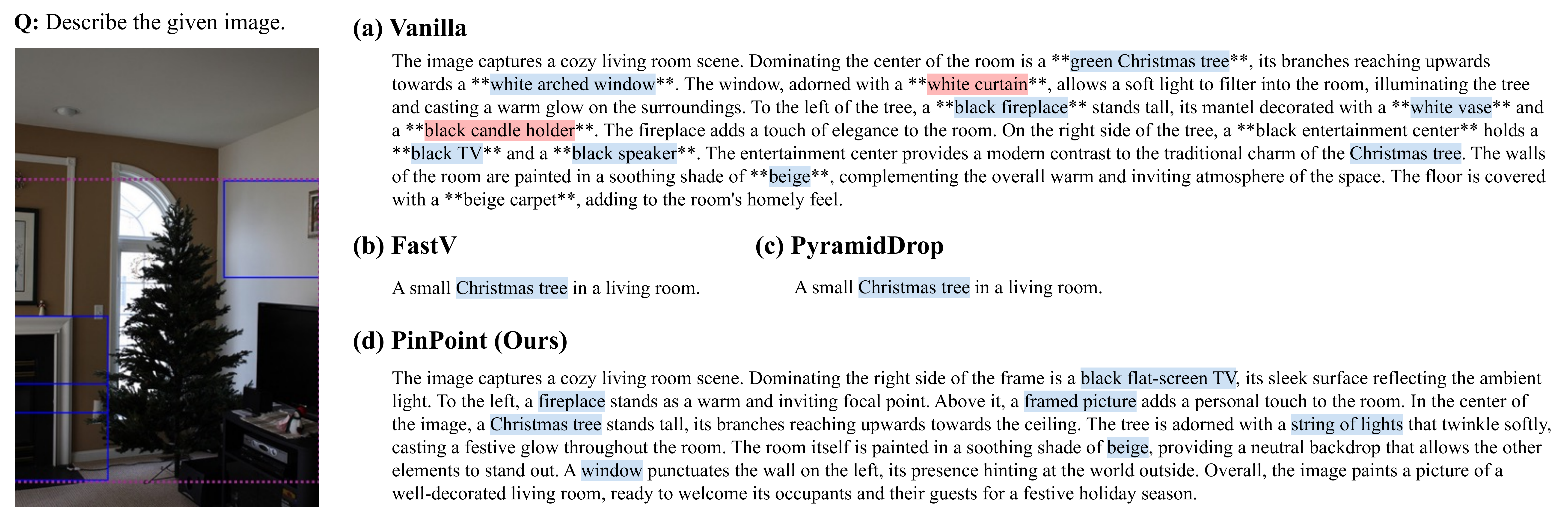}
  \caption{A qualitative comparison among LLaVA-NeXT~\cite{llavanext}, FastV~\cite{fastv}, PyramidDrop~\cite{pyramiddrop}, and our PinPoint shows that the vanilla model~\cite{llavanext} produces rich but often hallucinated descriptions, whereas FastV~\cite{fastv} and PyramidDrop~\cite{pyramiddrop} generate overly short outputs that miss key details. In contrast, PinPoint yields coherent, detailed descriptions with fewer hallucinations by leveraging instruction-relevant regions. Note that \textcolor{blue}{blue} marks the windows selected by PinPoint, and \textcolor{violet}{purple} indicates the pinpointed answer-relevant regions. In addition, \colorbox[rgb]{0.80, 0.88, 0.95}{light blue} indicates well-grounded keywords, whereas \colorbox[rgb]{0.99, 0.72, 0.72}{light red} denotes hallucinated keywords.}
  \label{fig:qualitative_captioning}
\end{figure*}
\begin{table}[h!]
\centering
\renewcommand{\arraystretch}{1}
\setlength{\tabcolsep}{1pt}
\scriptsize
\caption{\textbf{Performance comparison of training-based efficiency methods.} Our method outperforms Honeybee~\cite{honeybee} across all evaluated datasets with lower computational cost. Model: LLaVA-NeXT~\cite{llavanext}}
\label{tab:rebuttal_honeybee}
\resizebox{1.0\linewidth}{!}
{%
\begin{tabular}{
>{\raggedright\arraybackslash}m{1.4cm}|
>{\centering\arraybackslash}m{1.6cm}
>{\centering\arraybackslash}m{1.6cm}
>{\centering\arraybackslash}m{1.6cm}
>{\centering\arraybackslash}m{1.2cm}|
>{\centering\arraybackslash}m{1.6cm}
}
\toprule
\textbf{Method} 
& \textbf{InfoVQA}
& \textbf{SPDocVQA}
& \textbf{MPDocVQA}
& \textbf{GQA}
& \textbf{\shortstack{Avg.\\ FLOPs(T)}} \\
\midrule

Vanilla~\cite{llavanext}
& 0.2552 & 0.6628 & 0.3758 & 0.7598 & 42.77 \\

\midrule

Honeybee~\cite{honeybee}
& 0.2287 & 0.4451 & 0.2395 & 0.7460 & 27.80 \\

\rowcolor{gray!10} \textbf{Ours}
& \textbf{0.3024} & \textbf{0.6472} & \textbf{0.3866} & \textbf{0.7608} & 25.04 \\

\bottomrule
\end{tabular}
}

\end{table}

\begin{table*}[t!]
\centering
\renewcommand{\arraystretch}{1}
\setlength{\tabcolsep}{8pt}
\footnotesize
\caption{\textbf{Hallucination evaluation on POPE~\cite{pope} (discriminative) and CHAIR~\cite{chair} / AMBER~\cite{amber} (generative).} PinPoint achieves the lowest hallucination rates across POPE~\cite{pope}, CHAIR~\cite{chair}, and AMBER~\cite{amber}, outperforming in both discriminative and generative settings while maintaining strong reliability. Model: LLaVA-NeXT~\cite{llavanext}.}
\label{tab:hallucination}
{
\begin{tabular}{l|ccc|cc|cccc|c}
\toprule

\multirow{2}{*}{\textbf{\rule{0pt}{1.5em} Method}} 
& \multicolumn{3}{c|}{\textbf{POPE}}
& \multicolumn{2}{c|}{\textbf{CHAIR}}
& \multicolumn{4}{c|}{\textbf{AMBER}}
& \multirow{2}{*}{\textbf{\shortstack{\rule{0pt}{1em} Avg.\\ FLOPs(T)}}} \\ [2pt]

\cline{2-10}

& \rule{0pt}{1.2em} Rand. ↑
& Pop. ↑
& Adv. ↑
& CHAIR\textsubscript{S} ↓
& CHAIR\textsubscript{I} ↓
& CHAIR ↓
& Cover ↑
& Hal ↓
& Cog ↓
& \\

\midrule

Vanilla 
& 88.1 & 86.8 & 85.6
& 26.1 & 7.8
& 8.7 & \textbf{62.1} & 48.9 & 4.6
& 34.32 \\

\rowcolor{gray!10} \textbf{Ours}    
& \textbf{89.0} & \textbf{87.7} & \textbf{86.0}
& \textbf{25.6} & \textbf{7.1}
& \textbf{8.0} & 53.1 & \textbf{42.4} & \textbf{3.9}
& \textbf{22.18} \\

\bottomrule
\end{tabular}
}
\end{table*}

\begin{table*}[h!]
\centering
\renewcommand{\arraystretch}{1.1} 
\setlength{\tabcolsep}{6pt}      
\footnotesize
\caption{\textbf{Cross-dataset performance on general and real-world benchmarks.} Despite being trained exclusively on the GQA~\cite{gqa} dataset, PinPoint demonstrates superior generalization and robustness across challenging real-world tasks, including MMMU~\cite{mmmu}, MMMU-Pro~\cite{mmmu-pro} and TextVQA~\cite{textvqa}. Our method significantly outperforms state-of-the-art train-free approaches while achieving the lowest computational cost (FLOPs).}
\label{tab:rebutal_general}

\begin{tabular}{
>{\centering\arraybackslash}m{2.0cm} | 
>{\centering\arraybackslash}m{1.8cm} | 
>{\centering\arraybackslash}m{2.3cm} 
>{\centering\arraybackslash}m{2.4cm} 
>{\centering\arraybackslash}m{2.3cm} | 
>{\centering\arraybackslash}m{2.4cm} 
}
\toprule
\textbf{Method} 
& \textbf{Training Data} 
& \textbf{MMMU}
& \textbf{MMMU-Pro} (standard 10)
& \textbf{TextVQA}
& \textbf{Avg. FLOPs(T)} \\ 
\midrule

Vanilla~\cite{llavanext} 
& - & 0.3410 \scriptsize{(100.0\%)} & 0.1864 \scriptsize{(100.0\%)} & 0.5327 \scriptsize{(100.0\%)} & 39.92 \scriptsize{(100.0\%)} \\

\midrule
FastV~\cite{fastv}
& \textit{Train-free} & 0.3356 \scriptsize{(98.4\%)} & 0.1927 \scriptsize{(103.4\%)} & 0.5102 \scriptsize{(95.8\%)} & 28.17 \scriptsize{(70.6\%)} \\

PDrop~\cite{pyramiddrop}
& \textit{Train-free} & 0.3411 \scriptsize{(100.0\%)} & 0.1871 \scriptsize{(100.4\%)} & 0.5240 \scriptsize{(98.4\%)} & 26.00 \scriptsize{(65.1\%)} \\

\rowcolor{gray!10} \textbf{PinPoint (Ours)}
& \textbf{GQA} & \textbf{0.3500 \scriptsize{(102.6\%)}} & \textbf{0.1990 \scriptsize{(106.8\%)}} & \textbf{0.7293 \scriptsize{(136.9\%)}} & \textbf{24.39 \scriptsize{(61.1\%)}} \\

\bottomrule
\end{tabular}
\end{table*}








\myparagraph{Comparison with Training Method.}
To ensure a fair comparison, we evaluate our method against HoneyBee~\cite{honeybee}, a state-of-the-art approach that enhances VLM efficiency through projector training—specifically utilizing the C-Abstractor architecture. Under identical training configurations (e.g., data and epochs), PinPoint consistently outperforms HoneyBee in both efficiency (FLOPs) and performance (ANLS), as summarized in Table~\ref{tab:rebuttal_honeybee}. These results highlight our approach’s superiority in extracting instruction-relevant, fine-grained features without the accuracy trade-offs typically associated with conventional token reduction techniques.

\begin{table*}[t!]
\caption{\textbf{Effect of Encompass Supervision on Instruction–Region Alignment.} compare two supervision settings for instruction-relevant regions: 
(a) a single, tightly bounded answer box, and (b) encompass regions that aggregate multiple reasoning boxes covering all supporting visual evidence. 
Using encompass supervision (b) consistently yields higher ANLS and Region Accuracy across all datasets and base models, showing that including contextual supporting regions leads to better instruction–image alignment.}
\label{tab:dataset_comparison}
\centering
\renewcommand{\arraystretch}{1}
\setlength{\tabcolsep}{10pt}
\footnotesize
{
\begin{tabular}{l|c|cc|cc|cc}
\toprule

\multirow{2}{*}{\textbf{Model}}
& \multirow{2}{*}{\textbf{Setting}}
& \multicolumn{2}{c|}{\textbf{InfoVQA}}
& \multicolumn{2}{c|}{\textbf{SPDocVQA}}
& \multicolumn{2}{c}{\textbf{MPDocVQA}} \\

\cline{3-8}

\rule{0pt}{1.25em}& 
& ANLS$\uparrow$ & Region Acc.$\uparrow$
& ANLS$\uparrow$ & Region Acc.$\uparrow$
& ANLS$\uparrow$ & Region Acc.$\uparrow$ \\
\midrule

\multirow{2}{*}{LLaVA-NeXT-7B}
& (a) & 0.2980 & 82\% & 0.6457 & 97\% & 0.3836 & 86\% \\
& \cellcolor{gray!10} (b) & \cellcolor{gray!10} \textbf{0.3024} & \cellcolor{gray!10} \textbf{84\%} & \cellcolor{gray!10} \textbf{0.6472} & \cellcolor{gray!10} \textbf{98\%} & \cellcolor{gray!10} \textbf{0.3866} & \cellcolor{gray!10} \textbf{87\%} \\

\midrule

\multirow{2}{*}{Qwen2-VL-7B}
& (a) & 0.7017 & 93\% & 0.8843 & 96\% & 0.6572 & 90\% \\
& \cellcolor{gray!10} (b) & \cellcolor{gray!10} \textbf{0.7140} & \cellcolor{gray!10} \textbf{95\%} & \cellcolor{gray!10} \textbf{0.8977} & \cellcolor{gray!10} \textbf{98\%} & \cellcolor{gray!10} \textbf{0.6723} &  \cellcolor{gray!10} \textbf{94\%} \\

\bottomrule
\end{tabular}
}
\end{table*}

\begin{table*}[t!]
\caption{\textbf{Ablation on Region Coverage Threshold $\boldsymbol{r}$.}  We vary the region coverage threshold $r$ in the Region Selection stage and report answer accuracy (ANLS), region localization accuracy (Acc.), and region coverage (Cov.) for each dataset and base model. Across all settings, PinPoint maintains strong performance, with larger $r$ values increasing coverage and generally yielding modest ANLS gains.}
\label{tab:coverage_threshold}
\centering
\renewcommand{\arraystretch}{1}
\setlength{\tabcolsep}{7pt} 
\footnotesize
{
\begin{tabular}{l|c|ccc|ccc|ccc|ccc}
\toprule
\multirow{2}{*}{\textbf{Model}}  & \multirow{2}{*}{$\boldsymbol{r}$}  &
\multicolumn{3}{c|}{\textbf{InfoVQA}} &
\multicolumn{3}{c|}{\textbf{SPDocVQA}} &
\multicolumn{3}{c|}{\textbf{MPDocVQA}} &
\multicolumn{3}{c}{\textbf{GQA}} \\
\cline{3-14}

\rule{0pt}{1.25em}
& &
\textbf{ANLS} & \textbf{Acc.} & \textbf{Cov.} &
\textbf{ANLS} & \textbf{Acc.} & \textbf{Cov.} &
\textbf{ANLS} & \textbf{Acc.} & \textbf{Cov.} &
\textbf{ANLS} & \textbf{Acc.} & \textbf{Cov.} \\
\midrule

\multirow{3}{*}{LLaVA-NeXT-7B}
& 20\% & 0.2964 & 67\% & 36\% & 0.5724 & 83\% & 32\% & \textbf{0.3891} & 71\% & 36\% & 0.7219 & 85\% & 32\% \\
& 40\% & 0.3014 & 77\% & 55\% & 0.6270 & 93\% & 52\% & 0.3836 & 83\% & 56\% & 0.7508 & 93\% & 50\% \\
& \cellcolor{gray!10} 60\% & \cellcolor{gray!10} \textbf{0.3024} & \cellcolor{gray!10} 84\% & \cellcolor{gray!10} 71\% & \cellcolor{gray!10} \textbf{0.6472} & \cellcolor{gray!10} 98\% & \cellcolor{gray!10} 71\% & \cellcolor{gray!10} 0.3866 & \cellcolor{gray!10} 87\% & \cellcolor{gray!10} 72\% & \cellcolor{gray!10} \textbf{0.7608} & \cellcolor{gray!10} 98\% & \cellcolor{gray!10} 69\% \\
\midrule

\multirow{3}{*}{Qwen2-VL-7B}
& 20\% & 0.5882 & 69\% & 33\% & 0.7405 & 84\% & 30\% & 0.6094 & 73\% & 35\% & 0.7172 & 86\% & 30\% \\
& 40\% & 0.6649 & 87\% & 54\% & 0.8563 & 95\% & 54\% & 0.6578 & 86\% & 56\% & 0.7521 & 94\% & 51\% \\
& \cellcolor{gray!10} 60\% & \cellcolor{gray!10} \textbf{0.7140} & \cellcolor{gray!10} 95\% & \cellcolor{gray!10} 72\% & \cellcolor{gray!10} \textbf{0.8977} & \cellcolor{gray!10} 98\% & \cellcolor{gray!10} 72\% & \cellcolor{gray!10} \textbf{0.6723} & \cellcolor{gray!10} 94\% & \cellcolor{gray!10} 74\% & \cellcolor{gray!10} \textbf{0.7624} & \cellcolor{gray!10} 98\% & \cellcolor{gray!10} 71\% \\

\bottomrule
\end{tabular}
}
\vspace{-1.5em}
\end{table*}

\myparagraph{Evaluation on Hallucination Benchmark.} To evaluate the robustness of our method against hallucination, we conduct experiments on three established hallucination benchmarks that span both discriminative and generative settings. For POPE~\cite{pope}, a discriminative benchmark constructed on the MSCOCO dataset~\cite{mscoco}, we follow the standard evaluation protocol and report F1 scores across all categories. For the generative benchmarks CHAIR~\cite{chair} and AMBER~\cite{amber}, we evaluate on a representative subset by sampling 10\% of the MSCOCO~\cite{mscoco} validation split for CHAIR~\cite{mscoco} (approximately 4K frames) and follow the standard AMBER~\cite{amber} protocol. As shown in Table~\ref{tab:hallucination}, PinPoint consistently reduces hallucinated predictions across both discriminative and generative benchmarks, demonstrating strong reliability in mitigating hallucination and improving computational efficiency. We evaluate PinPoint trained on GQA~\cite{gqa}, since it shares a similar real-world image distribution with MSCOCO~\cite{mscoco}. The slight decrease in the Cover metric on AMBER~\cite{amber} is attributed to PinPoint’s focus on instruction-relevant regions rather than exhaustive object coverage, which aligns with our goal of grounding responses in the most semantically relevant visual evidence.

\myparagraph{Comparison with Cross-Data Setting.} To validate the generality and robustness of our approach, we evaluate PinPoint in a cross-dataset setting across expert-level reasoning tasks (MMMU~\cite{mmmu, mmmu-pro}) and real-world scene-text understanding (TextVQA~\cite{textvqa}). Despite being optimized exclusively on the GQA~\cite{gqa} dataset, PinPoint consistently surpasses all state-of-the-art train-free baselines. Notably, on TextVQA~\cite{textvqa}, it achieves a remarkable 136.9\% relative performance compared to the Vanilla baseline while reducing computational costs to only 61.1\% (24.39 TFLOPs). This significant performance-efficiency gain on unseen benchmarks demonstrates that PinPoint effectively masters the identification of instruction-relevant regions rather than overfitting to its training distribution. These results confirm that our module provides a robust, general-purpose solution for enhancing VLM efficiency without sacrificing complex reasoning capabilities in diverse real-world scenarios.

\subsection{Qualitative Results}
\myparagraph{Comparison for Caption Generation.} We qualitatively compare LLaVA-NeXT~\cite{llavanext}, FastV~\cite{fastv}, PyramidDrop~\cite{pyramiddrop}, and our proposed PinPoint on the image captioning task (i.e., describing the given image), as shown in Figure~\ref{fig:qualitative_captioning}. The vanilla LLaVA-NeXT~\cite{llavanext} generates rich captions but often introduces hallucinated content. In contrast, FastV~\cite{fastv} and PyramidDrop~\cite{pyramiddrop} produce overly brief responses that miss important visual details. PinPoint generates detailed and coherent captions by effectively leveraging instruction-relevant regions, while also reducing hallucinations.

\begin{figure*}[t!]
  \centering
  \includegraphics[width=1\linewidth]{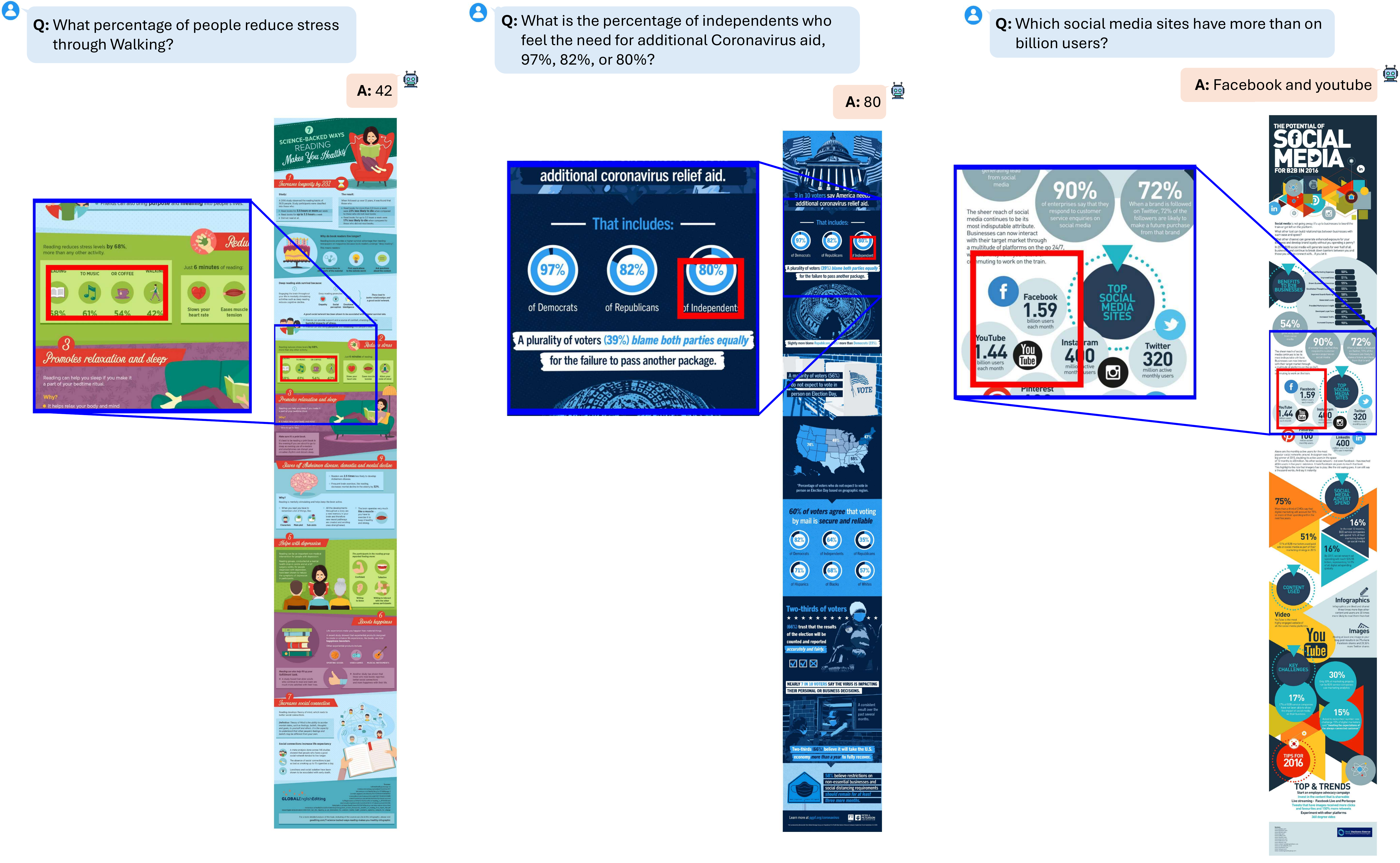}
  \caption{Additional Qualitative Results for PinPoint on InfoVQA~\cite{infographicvqa} large-size dataset. \textcolor{red}{Red} marks the ground-truth regions, \textcolor{blue}{blue} shows the window selected by our method.}
  \label{fig:qualitative_supp_2}
  \vspace{-0.5em}
\end{figure*}

\begin{figure*}[t!]
  \centering
  \includegraphics[width=1\linewidth]{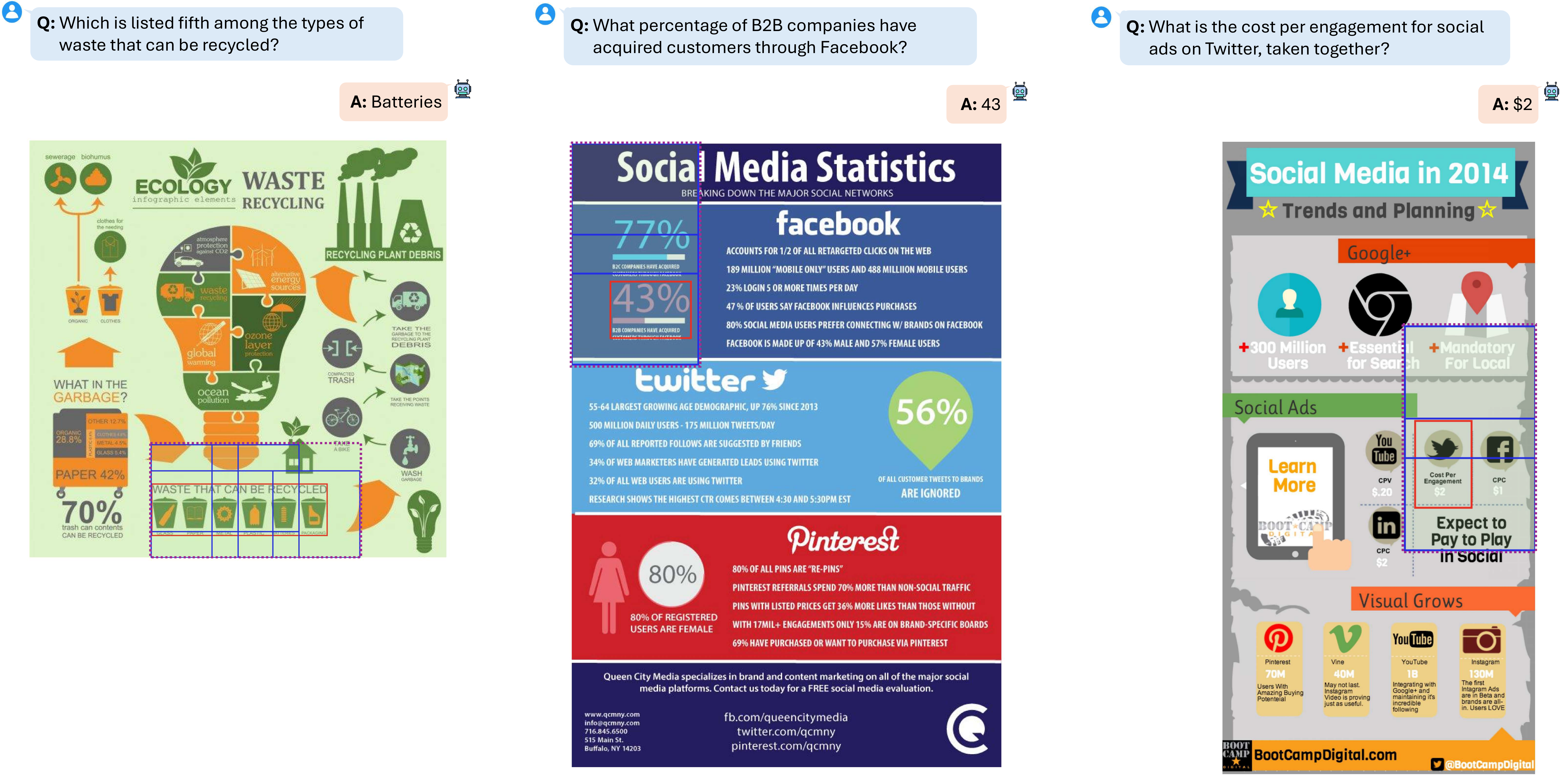}
  \caption{Additional Qualitative Results for PinPoint on InfoVQA~\cite{infographicvqa} medium-size dataset. \textcolor{red}{Red} marks the ground-truth regions, \textcolor{blue}{blue} shows the windows selected by our method, and \textcolor{violet}{purple} highlights the pinpointed answer-relevant areas.}
  \label{fig:qualitative_supp_3}
  \vspace{-0.5em}
\end{figure*}

\begin{figure*}[t!]
  \centering
  \includegraphics[width=1\linewidth]{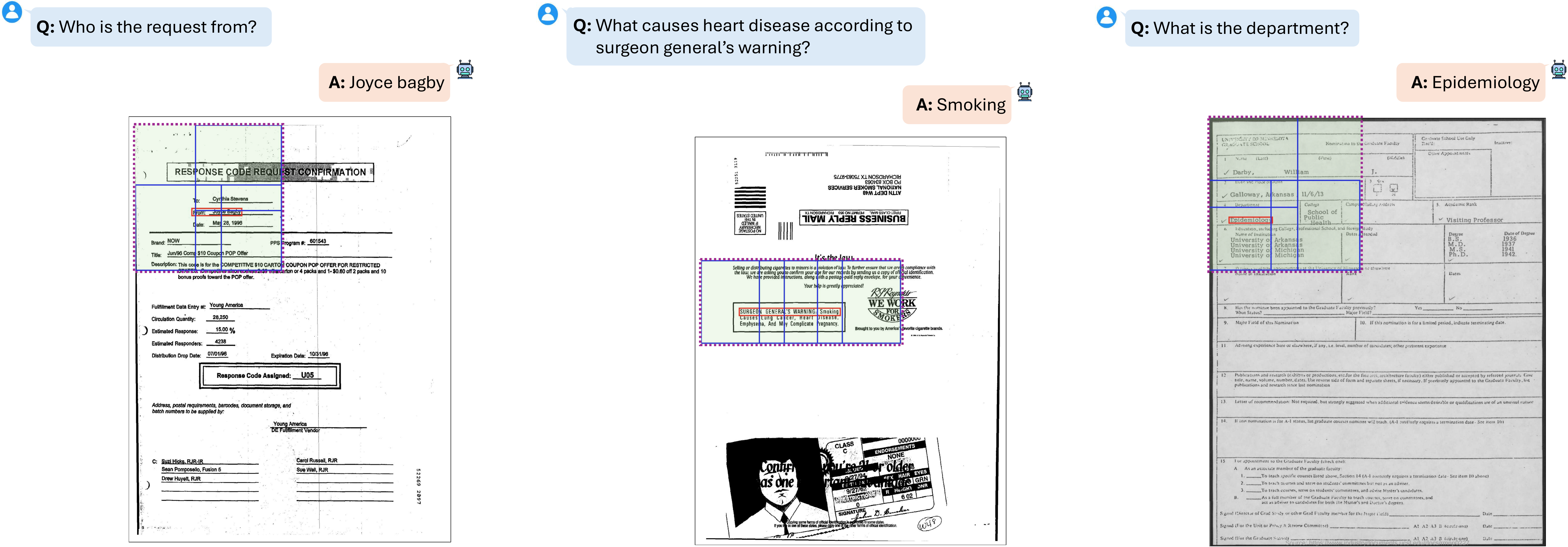}
  \caption{Additional Qualitative Results for PinPoint on SPDocVQA~\cite{docvqa} dataset. \textcolor{red}{Red} marks the ground-truth regions, \textcolor{blue}{blue} shows the windows selected by our method, and \textcolor{violet}{purple} highlights the pinpointed answer-relevant areas.}
  \label{fig:qualitative_supp_4}
  \vspace{-0.5em}
\end{figure*}

\begin{figure*}[!tp]
  \centering
  \includegraphics[width=1\linewidth]{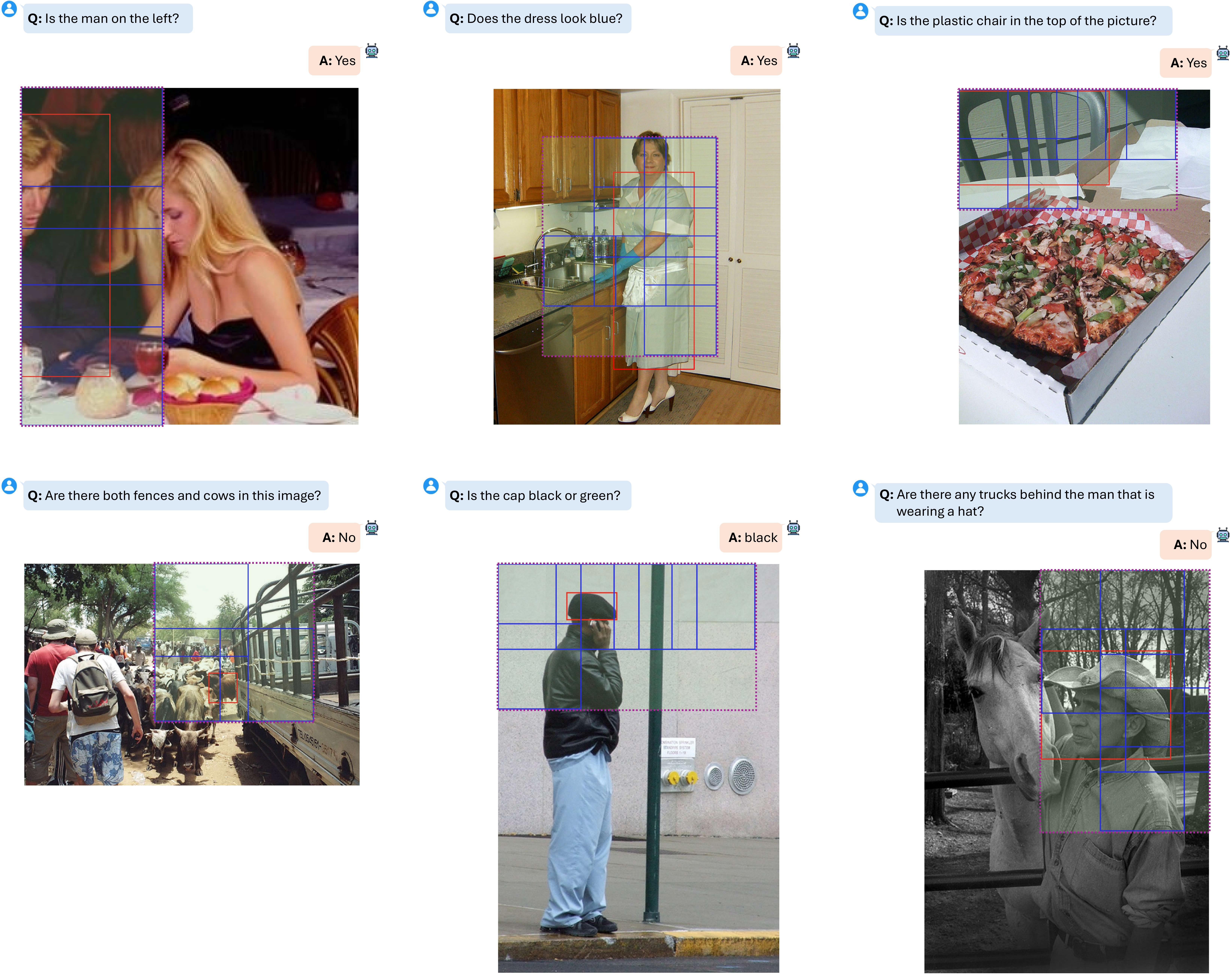}
  \caption{Additional Qualitative Results for PinPoint on GQA~\cite{gqa} dataset. \textcolor{red}{Red} marks the ground-truth regions, \textcolor{blue}{blue} shows the windows selected by our method, and \textcolor{violet}{purple} highlights the pinpointed answer-relevant areas.}
  \label{fig:qualitative_supp_1}
  \vspace{-0.5em}
\end{figure*}

\myparagraph{Additional Examples of PinPoint.} 
We provide additional qualitative examples showing that PinPoint reliably identifies instruction-relevant regions across diverse image sizes and visual layouts. These results indicate that our method consistently focuses on the appropriate visual evidence and produces accurate responses, regardless of variations in image resolution or scene complexity. Additional qualitative results are shown in Figure~\ref{fig:qualitative_supp_2} and~\ref{fig:qualitative_supp_3} for InfoVQA~\cite{infographicvqa}, in Figure~\ref{fig:qualitative_supp_4} for SPDocVQA~\cite{docvqa}, and in Figure~\ref{fig:qualitative_supp_1} for GQA~\cite{gqa}.

\renewcommand{\arraystretch}{1}
\setlength{\tabcolsep}{14pt}
\begin{table*}[t!]
\centering
\caption{\textbf{Number of Bounding Box Annotations and QA Pairs per Dataset.}}
\label{tab:dataset_stats}
\footnotesize
\begin{tabular}{l|l|ccc|c|c|c}
\toprule
\multirow{2}{*}{\textbf{Dataset}} & \multirow{2}{*}{\textbf{Split}} & \multicolumn{3}{c|}{\textbf{Annotation Count}} & \multirow{2}{*}{\textbf{\shortstack{Image \\Count}}} & \multirow{2}{*}{\textbf{\shortstack{Image Pages}}} & \multirow{2}{*}{\textbf{QA Pairs}} \\ \cline{3-5}
\rule{0pt}{1.2em}&& answer & evidence & encompass & & & \\ 
\midrule
\multirow{2}{*}{InfoVQA} & Train & 1.06 & 1.64 & 1.0& 4,162 & 1.0 & 17,887\\
& Validation & 1.09 & 1.53 & 1.0 & 500 & 1.0 & 2,801 \\
\midrule
\multirow{2}{*}{SPDocVQA}& Train & 1.00& 0.30 & 1.0 & 10,194& 1.0 & 39,463\\
& Validation & 1.00& 0.28 & 1.0& 1,286 & 1.0 & 5,349 \\
\midrule
\multirow{2}{*}{MPDocVQA} & Train & 1.00 & 1.05 & 1.0 & 5,131 & 9.1 & 36,230\\
& Validation & 1.00 & 1.02 & 1.0 & 927 & 5.6 & 5,187 \\ 
\bottomrule
\end{tabular}
\vspace{-1.5em}
\end{table*}
\begin{table}
\centering
\renewcommand{\arraystretch}{1}
\setlength{\tabcolsep}{9pt}
\footnotesize
\caption{\textbf{Distribution of InfoVQA~\cite{infographicvqa} Questions by Processing Type.}}
\label{tab:infovqa_percentage}
\begin{tabular}{l|cc}
\toprule
\textbf{Category} & \textbf{Train (\%)} & \textbf{Validation (\%)} \\
\midrule
Visually Grounded & 41.4 & 38.7 \\
OCR-Extractable (Unique) & 39.4 & 31.2 \\
OCR-Extractable (Multiple) & 12.6 & 21.9 \\
Manual Annotation & 6.6 & 8.2 \\
\bottomrule
\end{tabular}
\vspace{-.5em}
\end{table}

\begin{table}
\centering
\renewcommand{\arraystretch}{1}
\setlength{\tabcolsep}{9pt}
\footnotesize
\caption{\textbf{Distribution of SPDocVQA~\cite{docvqa} Questions by Processing Type.}}
\label{tab:spdoc_percentage}
\begin{tabular}{l|cc}
\toprule
\textbf{Category} & \textbf{Train (\%)} & \textbf{Validation (\%)} \\
\midrule
Visually Grounded & 35.2 & 28.0 \\
OCR-Extractable (Unique) & 62.6 & 59.0 \\
OCR-Extractable (Multiple) & 2.2 & 12.9 \\
Manual Annotation & 0.0 & 0.1 \\
\bottomrule
\end{tabular}
\vspace{-.5em}
\end{table}

\begin{table}
\centering
\renewcommand{\arraystretch}{1}
\setlength{\tabcolsep}{9pt}
\footnotesize
\caption{\textbf{Distribution of MPDocVQA~\cite{mpdocvqa} Questions by Processing Type.}}
\label{tab:mpdoc_percentage}
\begin{tabular}{l|cc}
\toprule
\textbf{Category} & \textbf{Train (\%)} & \textbf{Validation (\%)} \\
\midrule
Visually Grounded & 39.0 & 57.4 \\
OCR-Extractable (Unique) & 52.2 & 34.0 \\
OCR-Extractable (Multiple) & 8.6 & 7.9 \\
Manual Annotation & 0.2 & 0.7 \\
\midrule
\end{tabular}
\vspace{-1.5em}
\end{table}

\subsection{Additional Ablation}
\paragraph{Impact of Contextual Supporting Regions.}
\label{apdx:apdx_pinpoint_dataset}
We investigate how using datasets that explicitly include contextual regions around the answer influences instruction–region alignment. Our PinPoint dataset annotates not only the answer-containing region but also an encompass region that includes surrounding contextual elements necessary to infer the answer. As shown in Table~\ref{tab:dataset_comparison}, training with only the tight answer box (setting (a)) yields consistently lower Region Accuracy and ANLS than training with encompass annotations that cover all supporting evidence (setting (b)). This suggests that effective cross-modal alignment benefits from supervising not just the answer location itself, but the broader set of answer-related regions that ground the instruction semantically. 

\myparagraph{Sensitivity to the Region Coverage Threshold $\boldsymbol{r}$.}
In the Region Selection stage, the number of selected regions $k$ is adaptively determined by a predefined region coverage threshold $r$. Table~\ref{tab:coverage_threshold} reports, for each value of $r$, the resulting region coverage (Cov.), region localization accuracy (Acc.), and ANLS of the generated answers. When using LLaVA-NeXT~\cite{llavanext} as the base model, performance on fine-grained datasets that require focusing on small details (e.g., InfoVQA, MPDocVQA) remains relatively stable as $r$ varies, whereas datasets that involve broader layouts (e.g., SPDocVQA, GQA) benefit from larger coverage and show improved ANLS with higher $r$. For Qwen2-VL~\cite{qwen2vl}, enlarging the covered area generally improves performance, indicating that this model gains from seeing more surrounding context. Notably, even at $r=40\%$, where only about 40\% of the image area is retained, PinPoint already achieves strong accuracy compared to competing methods (see main results), suggesting that our instruction–region alignment remains effective over a wide range of coverage thresholds.

\section{Dataset}
\label{apdx:apdx_dataset}
\subsection{Dataset Generation Pipeline}
\label{apdx:apdx_datapipeline}
We construct new annotated datasets for InfographicVQA (InfoVQA)~\cite{infographicvqa}, SinglePageDocVQA (SPDocVQA)~\cite{docvqa}, and MultiPageDocVQA (MPDocVQA)~\cite{mpdocvqa}, in which we explicitly mark regions that are relevant to answering the question. These annotations go beyond a single answer box and include essential supporting evidence. For instance, for the question “What is the object the man is pointing at?”, it is necessary not only to localize the pointed object but also to first locate the man; our annotations therefore include both the man and the target object as instruction-relevant regions.

To obtain these annotations, we design a unified dataset construction pipeline that is shared across all benchmarks, with minor adjustments to accommodate dataset-specific characteristics. The pipeline combines (i) an OCR engine (Amazon Textract), (ii) a vision-language model (Qwen2.5-VL~\cite{qwen25vl}), and (iii) a large language model (GPT-4o~\cite{gpt4}).

Relying on a single model to directly predict answer regions is often unreliable for information-dense (e.g., document layout) images, so we handle each sample using three complementary cases based on the OCR outcome. If the answer-related text can be found by OCR and appears exactly once, we directly use the corresponding bounding box as the instruction-relevant region. If multiple candidate boxes contain the answer text, we retain all candidates and use an large-language model to perform additional reasoning given the question, answer, OCR text, and candidate regions, selecting only the most relevant box. If the answer text cannot be recovered by OCR, indicating that complex visual reasoning is required, we instead use a vision-language model to infer and annotate the appropriate answer-related region.


While the overall pipeline structure is consistent, we introduced a specific enhancement for the challenging InfoVQA~\cite{infographicvqa} dataset, which demands complex reasoning. For this dataset, an LLM generates rationale sentences from the input informations (question, answer, and OCR) to guide the localization of the corresponding reasoning-related regions. Finally, all datasets underwent a quality control step where instances with failed or invalid bounding box outputs were manually corrected via self-annotation, ensuring high-quality supervision. 

\subsection{Dataset Analysis}
\label{apdx:apdx_dataanalysis}

Table~\ref{tab:dataset_stats} summarizes the detailed statistics of our newly constructed annotations. For InfoVQA~\cite{infographicvqa}, the most challenging benchmark, answers often appear at multiple locations, and each question typically requires the largest number of evidence regions to be correctly resolved. Although both SPDocVQA~\cite{docvqa} and MPDocVQA~\cite{mpdocvqa} are document-based datasets, MPDocVQA~\cite{mpdocvqa} consistently demands more supporting evidence per question, indicating a higher level of compositional and cross-page reasoning.

As described in Section~\ref{apdx:apdx_datapipeline}, our pipeline processes each sample through four dedicated cases to accurately identify answer-relevant regions. For InfoVQA~\cite{infographicvqa}, Table~\ref{tab:infovqa_percentage} shows that the Visually Grounded case accounts for nearly 40\% of all samples, indicating that a large portion of questions cannot be resolved from OCR text alone and genuinely require visual reasoning. In contrast, for the document-layout datasets SPDocVQA~\cite{docvqa} and MPDocVQA~\cite{mpdocvqa}, Tables~\ref{tab:spdoc_percentage} and~\ref{tab:mpdoc_percentage} shows that OCR-based cases constitute the majority, reflecting that most answers can be localized primarily via textual cues.

Across all datasets, a substantial fraction of samples still relies on joint text–image reasoning, which supports our decision to explicitly separate and handle these cases within the pipeline. Moreover, the proportion of manually annotated samples remains below 10\% on every benchmark, demonstrating that our method significantly reduces human labeling effort while maintaining high annotation quality.

\subsection{Dataset Examples}
\begin{figure*}[t!]
  \centering
  \includegraphics[width=1\linewidth]{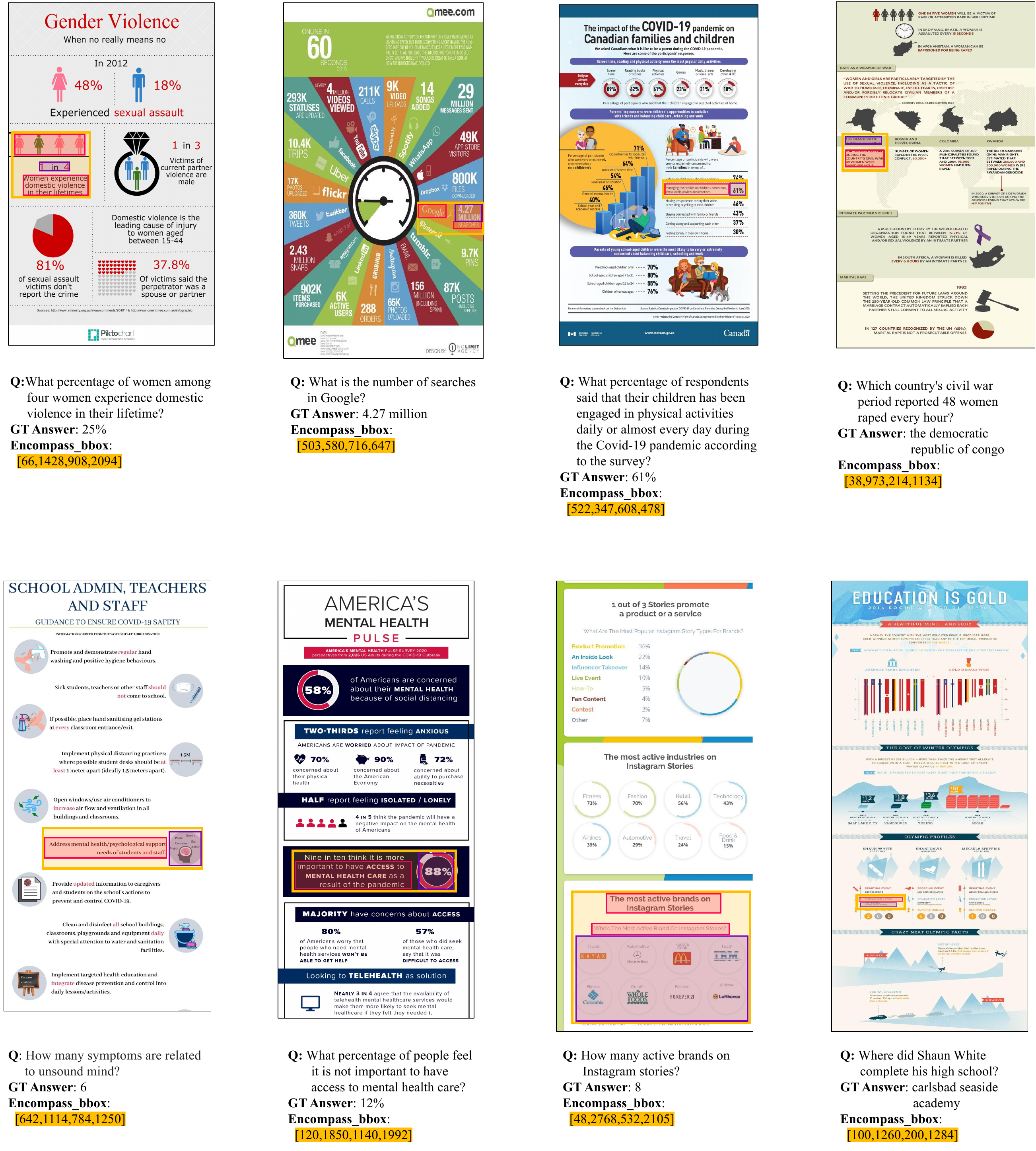}
  \caption{Qualitative examples of the PinPoint annotations on InfoVQA~\cite{infographicvqa}. \textcolor{violet}{Purple} boxes denote answer regions, \textcolor{magenta}{pink} boxes indicate supporting evidence regions, and \textcolor{orange}{yellow} boxes represent encompass regions that jointly cover both answer and evidence, providing a more complete instruction-relevant area for supervision.}
  \label{fig:qualitative_supp_data_1}
  \vspace{-0.5em}
\end{figure*}

\begin{figure*}[t!]
  \centering
  \includegraphics[width=1\linewidth]{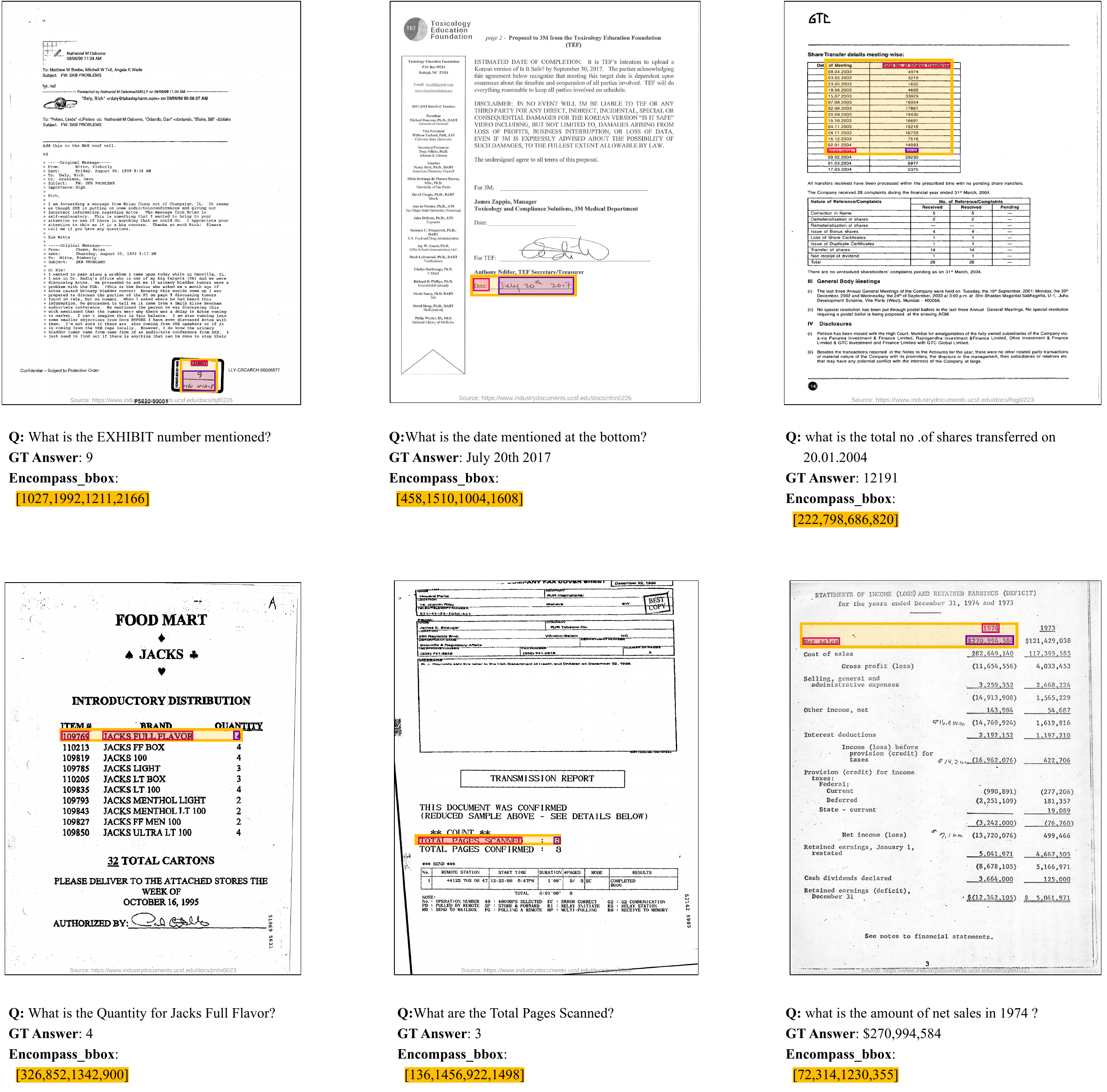}
  \caption{Qualitative examples of the PinPoint annotations on SPDocVQA~\cite{docvqa}. \textcolor{violet}{Purple} boxes denote answer regions, \textcolor{magenta}{pink} boxes indicate supporting evidence regions, and \textcolor{orange}{yellow} boxes represent encompass regions that jointly cover both answer and evidence, providing a more complete instruction-relevant area for supervision.}
  \label{fig:qualitative_supp_data_2}
  \vspace{-0.5em}
\end{figure*}

Figure~\ref{fig:qualitative_supp_data_1} and~\ref{fig:qualitative_supp_data_2} present qualitative examples of the annotations on InfoVQA~\cite{infographicvqa} and SPDocVQA~\cite{docvqa}. The encompass regions unify the answer and its supporting evidence into a single region, allowing the model to learn complete instruction-relevant cues. These examples illustrate that a substantial portion of document-understanding questions require multi-step reasoning across spatially dispersed content, and the encompass annotations effectively capture all necessary components for deriving the correct answer.

\subsection{Prompt Design}
\label{apdx:apdx_prompt}
\begin{figure*}[h!]
  \centering
  \includegraphics[width=1.0\linewidth]{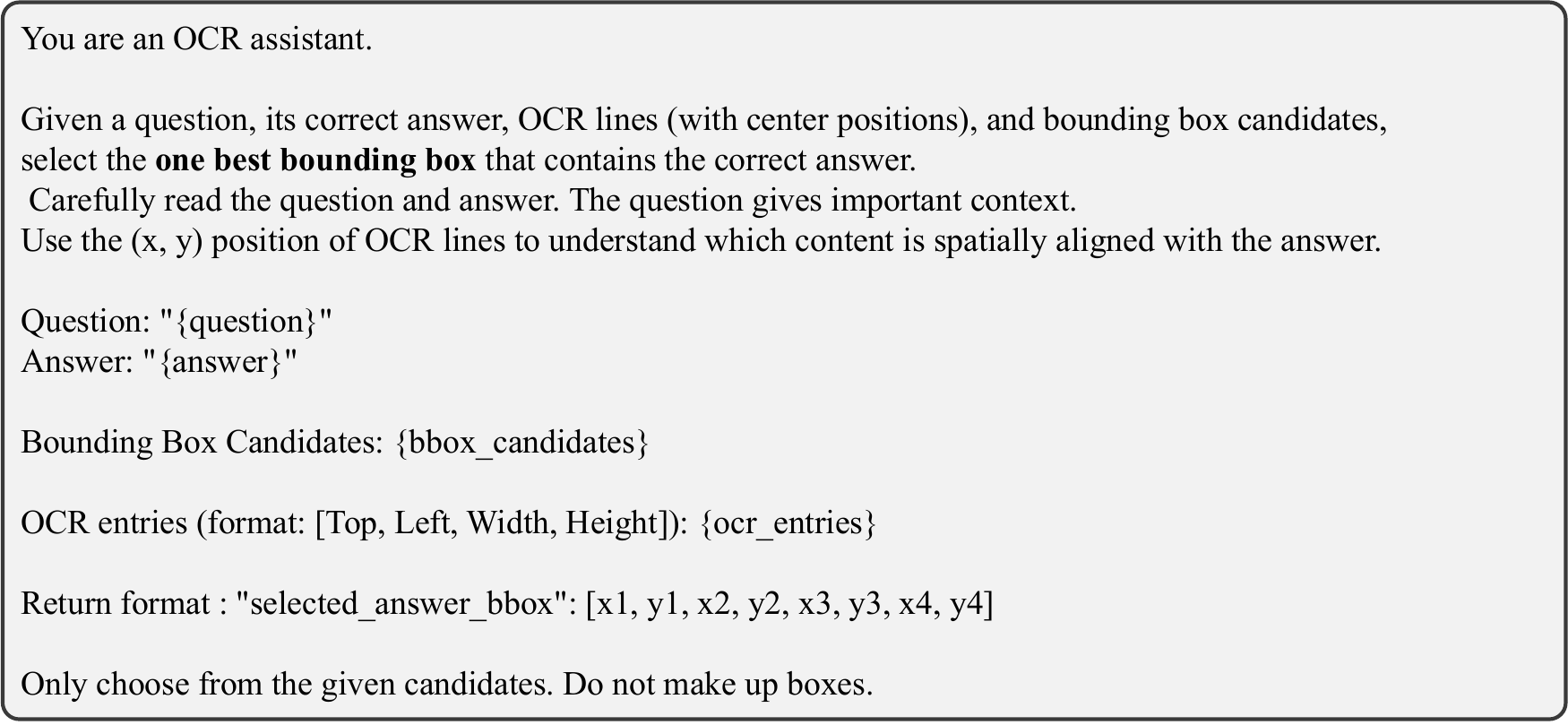}
  \caption{Prompt for selecting a single bounding box among multiple candidates.}
  \label{afig:prompt_1}
\end{figure*}

\begin{figure*}[h!]
  \centering
  \includegraphics[width=1.0\linewidth]{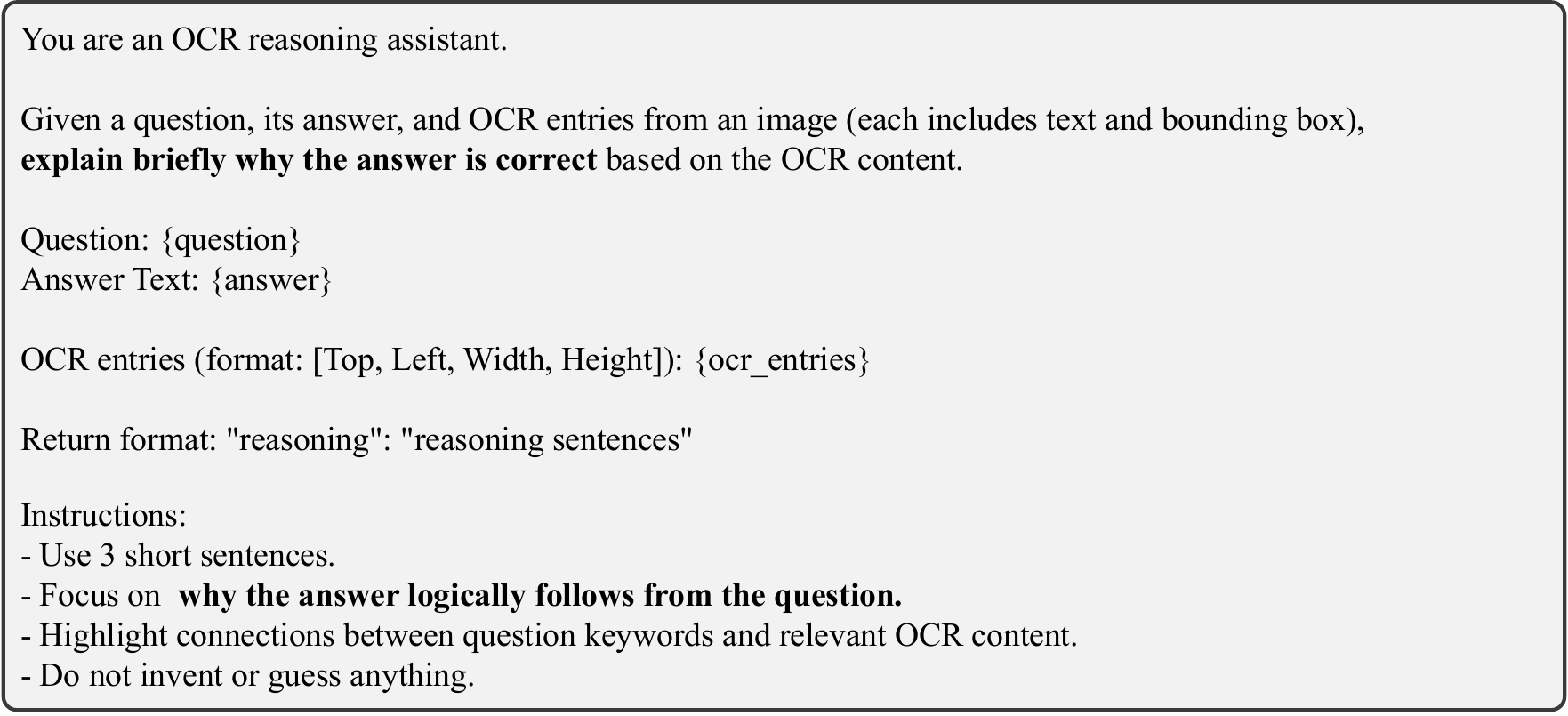}
  \caption{prompt for generating a reasoning sentence required to derive the correct answer.}
  \label{afig:prompt_2}
\end{figure*}

\begin{figure*}[h!]
  \centering
  \includegraphics[width=1.0\linewidth]{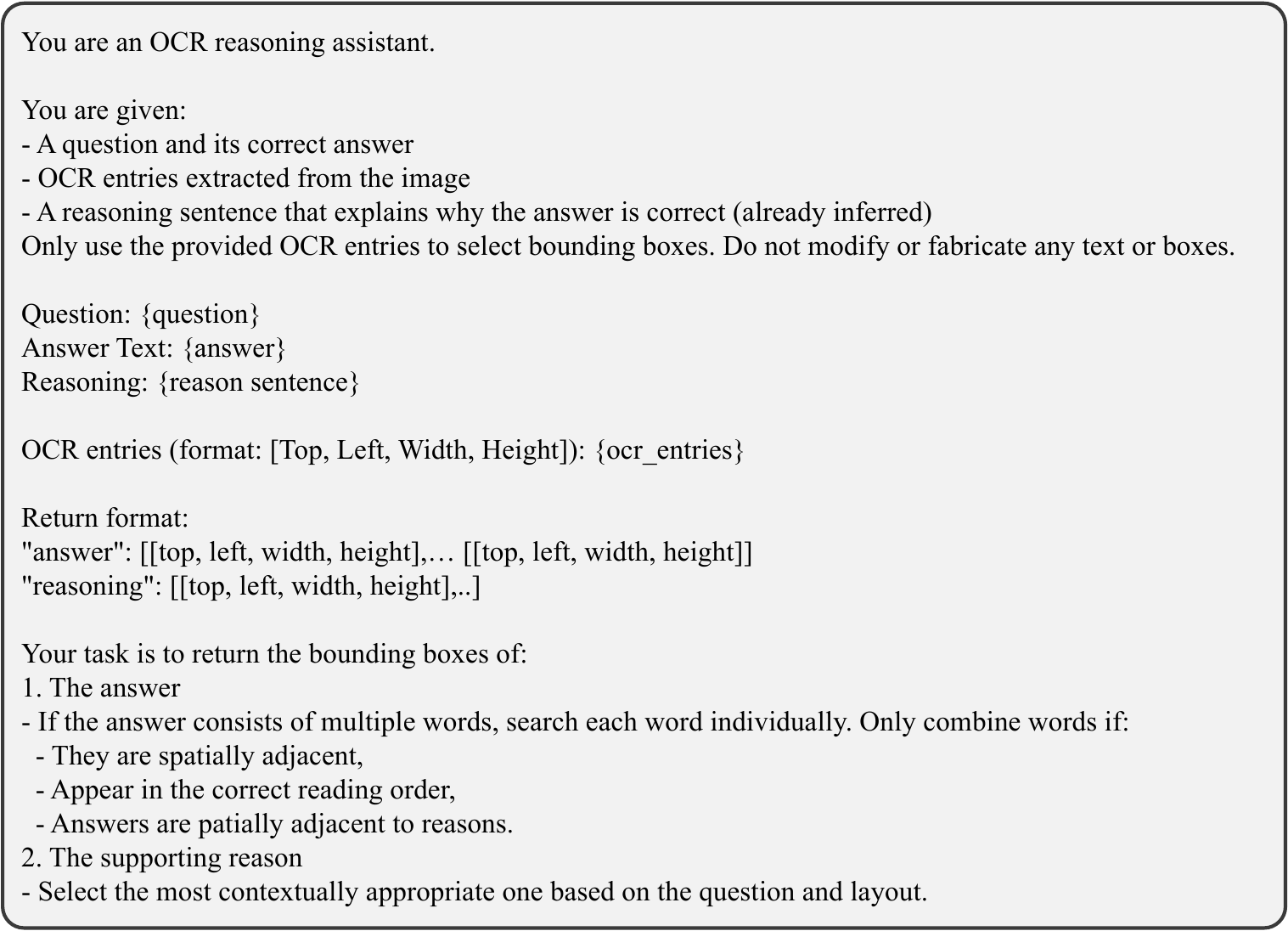}
  \caption{Prompt for extracting and localizing the answer and supporting reasoning elements from OCR entries.}
  \label{afig:prompt_3}
\end{figure*}

\begin{figure*}[h!]
  \centering
  \includegraphics[width=1.0\linewidth]{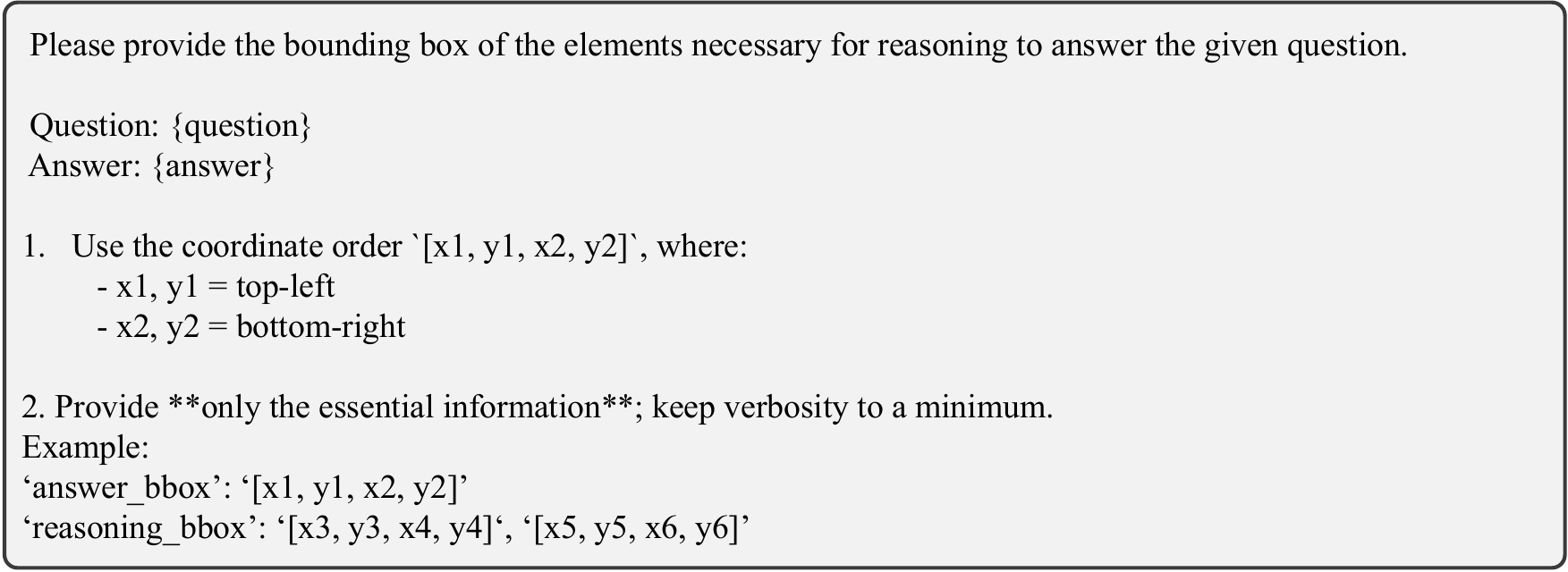}
  \caption{Prompt for identifying instruction-relevant regions using a VLM}
  \label{afig:prompt_4}
\end{figure*}

We next describe the prompts used to construct the PinPoint dataset annotations. Figure~\ref{afig:prompt_1} illustrates the prompt used when multiple candidate bounding boxes contain the answer text, where an LLM is asked to select the region most relevant to the question. Figures~\ref{afig:prompt_2} and~\ref{afig:prompt_3} show the prompts additionally used for InfoVQA~\cite{infographicvqa}: the first elicits rationale sentences given the question, answer, and OCR text, and the second leverages these rationales to localize the corresponding reasoning-related regions. Finally, Figure~\ref{afig:prompt_4} presents the prompt used when OCR fails to recover the answer text, in which a vision-language model performs visual reasoning to identify and annotate the answer-related region directly from the image.

\clearpage



\clearpage


\end{document}